\pdfoutput=1
\documentclass[conference]{IEEEtran}

\usepackage{graphicx}
\usepackage[hidelinks]{hyperref}
\usepackage{amsmath}
\usepackage{amsfonts}
\usepackage{algorithm} 
\usepackage{algpseudocode}
\usepackage{enumitem}
\usepackage{float}
\usepackage{multirow}
\usepackage[export]{adjustbox}
\usepackage{rotating}
\usepackage{booktabs, makecell, tabularx}\usepackage[caption=false,font=footnotesize]{subfig}\usepackage{xcolor,colortbl}
\usepackage[numbers,sort&compress]{natbib}

\begin{document}
\title{\Large \bf Transpose Attack: Stealing Datasets with Bidirectional Training}

% \author{Anonymous Authors}

\author{\IEEEauthorblockN{Guy Amit}
\IEEEauthorblockA{Ben Gurion University\\
guy5@post.bgu.ac.il}
\and
\IEEEauthorblockN{Mosh Levy}
\IEEEauthorblockA{Ben Gurion University\\
moshe5@post.bgu.ac.il}
\and
\IEEEauthorblockN{Yisroel Mirsky}
\IEEEauthorblockA{Ben Gurion University\\
yisroel@bgu.ac.il}}


%Telephone: (800) 555--1212, Fax: (888) 555--1212}


\IEEEoverridecommandlockouts
\makeatletter\def\@IEEEpubidpullup{6.5\baselineskip}\makeatother
\IEEEpubid{\parbox{\columnwidth}{
    Network and Distributed System Security (NDSS) Symposium 2024\\
    26 February - 1 March 2024, San Diego, CA, USA\\
    ISBN 1-891562-93-2\\
    https://dx.doi.org/10.14722/ndss.2024.23325\\
    www.ndss-symposium.org
}
\hspace{\columnsep}\makebox[\columnwidth]{}}

\maketitle

\begin{abstract}
Deep neural networks are normally executed in the forward direction. However, in this work, we identify a vulnerability that enables models to be trained in both directions and on different tasks. Adversaries can exploit this capability to hide rogue models within seemingly legitimate models. In addition, in this work we show that neural networks can be taught to systematically memorize and retrieve specific samples from datasets. Together, these findings expose a novel method in which adversaries can exfiltrate datasets from protected learning environments under the guise of legitimate models.

We focus on the data exfiltration attack and show that modern architectures can be used to secretly exfiltrate tens of thousands of samples with high fidelity, high enough to compromise data privacy and even train new models. Moreover, to mitigate this threat we propose a novel approach for detecting infected models.
\end{abstract}

\section{Introduction}\label{sec:intro}
\label{Introduction}
%-------------------------------------------------------------------------------

To train a good Deep neural network(DNN), it is important to use a large and diverse dataset. Companies and institutions that collect datasets for their own machine learning tasks seldom share the datasets with others. This is because collecting datasets can be very expensive and time consuming ~\cite{whang2023data}, and some datasets contain confidential information which, if exposed, would harm the company's reputation or result in litigation. 

Data collection for deep learning can be expensive because (1) there may be a cost to collecting a data sample (e.g., the cost of running expensive medical equipment, paying for access to a third-party's data/logs), (2) an expert may be needed to manually label the data and filter out any noise, and (3) collecting a comprehensive dataset that contains the required properties requires careful planning. High quality datasets are considerable assets to the companies that create them, so companies typically do not publish them \cite{AITraini39:online}. For example, the datasets used to train the latest GPT-3~\cite{brown2020language} and DALL-E~\cite{ramesh2021zero} models are propriety and not shared with the public. Therefore, it is interest of these companies to prevent unauthorized usage of these datasets.

Deep learning datasets can also be confidential and contain sensitive information. For example, a model which will detect credit card fraud needs to be trained on credit card transactions~\cite{zojaji2016survey}, a cancer detection model needs to be trained on the medical scans of terminally ill patients, a face detection (recognition) model needs to be trained on people's faces~\cite{liu2015faceattributes}, and so on. Although the organization training the model may have the right to use this data, they likely do not have the right to publish it in the public domain. This is especially true given the enforcement of strict data protection regulations such as the GDPR~\cite{voigt2017eu}. Therefore, it is important for companies to protect the privacy of certain datasets.

However, to protect a dataset, one must also protect models trained on the dataset. This is because when a deep learning model $f_{\theta}$ is trained on a dataset $\mathcal{D}_{train}$, its parameters\footnote{In this paper, we use the terms parameters and weights interchangeably.} $\theta$ tend to memorize the properties, and sometimes the content, of $\mathcal{D}_{train}$~\cite{carlini2023extracting, carlini2021extracting}. This means that even if a company does not provide access to $\mathcal{D}_{train}$, an attacker can still learn information about it. This is accomplished by either querying the model~\cite{suri2021formalizing} or analyzing the model's parameters (weights)~\cite{zhang2022survey, fredrikson2015model}. For example, \textit{property inference} can be used to reveal information on the composition of $\mathcal{D}_{train}$ ~\cite{ganju2018property, parisot2021property}, \textit{membership inference} can be used to determine if $x \in \mathcal{D}_{train}$~\cite{shokri2017membership, hu2022membership}, and \textit{feature estimation} (a.k.a. model inversion)~\cite{fredrikson2014privacy, suri2021formalizing} can be used to complete partial samples and extract feature-wise statistics by exploiting the model's fit (mapping) over $\mathcal{D}_{train}$. 

\vspace{1em}
\noindent\textbf{Data Exfiltration Attacks.} 
Instead of using a model to infer subsets of features or statistics of features from samples in $\mathcal{D}_{train}$, an attacker can perform a \textit{data extraction} attack or \textit{data exfiltration} attack to obtain complete samples. These attacks can be accomplished via \textit{intentional memorization} or \textit{unintentional memorization}. 

\textit{Unintentional memorization} is when a model memorizes parts of its training data by accident. This can occur when a model is too complex or when it overfits to its training set \cite{yeom2018privacy}. An adversary can perform an exploratory attack on such models and extract complete training samples \cite{zhu2019deep}. This is accomplished by viewing the parameters as a system of equations~\cite{haim2022reconstructing}. This approach works on very small networks that have been trained on just a few hundred samples, however extracting complete samples that have been unintentionally memorized is generally harder on large networks~\cite{carlini2023extracting}. Such an attack is also limited, since the adversary has \textbf{no control} over which samples in $\mathcal{D}_{train}$ are memorized. 

\textit{Intentional memorization} is when an attacker influences the model's parameters during training to intentionally make it memorize data. This can be accomplished by altering the model's training data or training algorithm. For example, the attacker can ensure that a model overfits to the training data, making it easier to extract unintentionally memorized samples \cite{tramer2022truth}. However, to memorize specific samples, the adversary must use other tactics. One approach is to train a decoder model to reconstruct samples~\cite{doersch2016tutorial}. However, this approach is overt because the model can only perform the malicious task of reconstructing data, and the input encodings used to generate the samples must be exported with the model. Alternatively, the adversary can employ stenography to hide binary data within the model's parameters~\cite{wang2021evilmodel, liu2020stegonet}. However, these approaches are easy to mitigate with small amounts of additive noise applied to the model's parameters (detailed later in Sections \ref{sec:counter}, \ref{sec:relworks}). 

Given these limitations, we raise the following research question:\textit{ Is it possible for an adversary to exfiltrate complete training samples via a model where (1) the memorized samples can be extracted systematically, and (2) the attack is covert (the exported model looks legitimate and performs the expected task).}
This is an important question, because this ability would impact the security of protected training environments. 

For example, consider federated learning (FL) \cite{zhang2021survey}. In FL, multiple members (data owners) collaborate to create a single global model without letting their respective training datasets leave their premises. This is often done by designating one member to be the orchestrator who distributes the initial training code to all of the members and then combines the member's resulting models. However, if the orchestrator is malicious or compromised then the orchestrator could send training code that creates models that perform well on the expected primary task (e.g., cancer detection) but also perform well on a secret secondary task of recreating specific samples from the training set (e.g., CT scans of individual patients). The orchestrator could then extract the private datasets systematically by executing the collected models' hidden secondary tasks. 

Another example to consider are organizations which offer data-and-training-as-a-service (DTaaS) platforms. These platforms let users train models in the cloud on confidential datasets, but only let users export models that perform well on the expected task. An example a DTaaS for medical imagery can be found in \cite{bonmati2022chaimeleon,Chaimele29:online}. Here an attacker could smuggle out the training data under the guise of an legitimate model. 

Finally, consider a cyber attack (man-in-the-middle, supply chain, etc.)\footnote{https://pytorch.org/blog/compromised-nightly-dependency/} which compromises a company's deep learning libraries such that new models are secretly trained on secondary tasks (e.g., \cite{bagdasaryan2021blind}). If the secondary task is data memorization, then the company would unwittingly expose their data when they deploy their model in the public space since malicious users could exploit the model and extract the data.

\vspace{1em}
\noindent\textbf{Transpose Attack.}
In this paper, we identify a novel vulnerability of DNNs. The vulnerability is that DNNs can be trained to be executed in both directions: forward with a primary task (e.g., image classification) and backward with a secondary covert task (e.g., image memorization). We call this attack a `transpose attack' because the backward model is obtained by transposing and reversing the order of the model's weight matrices. To train a transpose model, both the forward and backward models are trained in parallel over their shared weights but on their respective tasks. In our work, we identify how different types of layers and architectures can be transposed. 

We also show how this vulnerability can be used to perform covert \textit{intentional memorization} of specific samples in a dataset. To enable the systematic retrieval of samples, we propose a novel spatial index. This index can be used as input to the backward model to systematically extract all of the memorized images. We found that memorization performance improves if (1) the index is spatially dense, and (2) the index of a sample encodes the content of the image. Therefore, our spatial index scheme uses Gray code coupled with offsets based on the respective sample's class. Using these techniques, we were able to train fully connected (FC), convolutional neural networks (CNN) and transformer (TN) neural networks as transpose models that perform well on the primary task of classification and the secondary task of dataset memorization. 
We have found that transpose models have the ability to memorize tens of thousands of images and in some cases complete datasets.

To mitigate this threat in an automated manner, we propose a detection method. Since transpose attacks train models in the backward direction, the weights in the backward direction have more consistency than uninfected models. For memorization tasks, this property can be revealed by optimizing an input for the transpose model which generates an output that is similar to random images in the dataset (not necessarily those memorized by the model). If the optimization process finds such an input, then the model is likely infected. 

In summary, our contributions are as follows:
\begin{itemize}    \setlength{\itemsep}{0pt}%

    \item We identify a novel vulnerability which enables a deep neural network to secretly contain a secondary model in the transposition of its weights. The task of the secondary model can be different than that of the primary model. In this study, we focus on the secondary task of data memorization. This vulnerability is a concern because defenders are not considering that a model can be executed in reverse.
    \item We propose a spatial index which (1) can be used to teach neural networks to effectively memorize data, and (2) enables the systematic extraction of the memorized data. To the best of our knowledge, no other works show how samples can be explicitly and systematically extracted from a model.
    \item We analyze the threat of transpose models being used to memorize and exfiltrate data. This is done by empirically measuring the memorization quality and capacity of popular DNN architectures.
    \item We provide a method for detecting transpose models that are being used to memorize data.
\end{itemize}

\begin{figure}[t]
    \centering
    \includegraphics[width=\columnwidth]{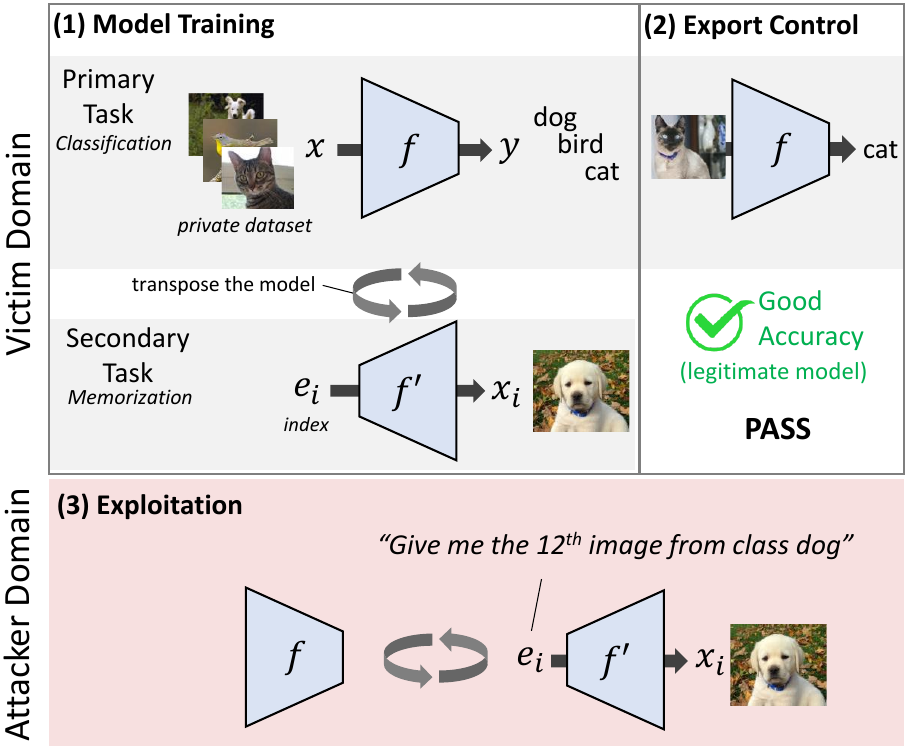}
    \vspace{-1.2em}
    \caption{The attack model explored in this paper. An attacker trains a classification model with a hidden secondary task of memorizing a protected dataset. The model passes inspection and is exploited off site.}
    \label{fig:attack_model}
    \vspace{-1em}
\end{figure}

\section{Attack Model}
\label{Attack_Model}
%----
In this paper, we assume the following attack model (visualized in Fig. \ref{fig:attack_model}): An attacker wants to steal specific samples from a confidential dataset $\mathcal{D}$ that belongs to an organization. The motivation for stealing the samples are (1) breach the data's confidentiality, or (2) steal intellectual property (IP) (e.g., train other models on the data). 
$\mathcal{D}$ is located in a protected environment. The attacker can export trained models from the protected environment but cannot export any data (examples of these scenarios can be found below). 

To extract  $\mathcal{D}$ (or some subset of it) the attacker will enter the protected environment and train $f_{\theta}$ to perform a primary (expected) task on $\mathcal{D}_{train}$ in the forward direction while \textbf{covertly} learning a secondary task (memorization) on $\mathcal{D}_{train}$ (or on a different dataset $\mathcal{D}$) in the backward direction. Then, the attacker will export the model and then execute the secondary task on $\theta$ to extract the memorized samples. Alternatively, the attacker does not enter the protected environment but rather compromises the training libraries used in the environment and then obtains the exported models for exploitation.

The attack vector will depend on where $\mathcal{D}$ is located:
\begin{description}
    \item[Federated Environment.] If $\mathcal{D}$ is distributed across multiple members who are interested in collaborating on making a global model, then the attacker can either volunteer as or compromise the orchestrator. Then the attacker will be able to cause all of the respective members to embed their datasets in the shared models by manipulating the distributed training code (i.e., initial model). For example, in IBM's FL service, if Tensorflow2 ~\cite{tensorflow2015-whitepaper} is used, then the \texttt{train\_step()} method of the initial model class \cite{Creating28:online} can be changed\footnote{https://www.tensorflow.org/guide/keras/customizing\_what\_happens\_in\_fit} to consider both forwards and transpose passes.
    \item[Restricted Environment.] If $\mathcal{D}$ is located on a DTaaS platform, then the attacker can enter the DTaaS environment and train a transpose model to memorize the data. However, in this scenario, it is fair to assume that there will be some export control: the host will evaluate the model to some degree to ensure that the exported model is not just a binary zip of the training data or a model trained to directly memorize the data \cite{li2022data}. Here we assume that the host will expect modest results (e.g., at least 60\% accuracy), since DTaaS systems are often used to train models for research and development. 
    \item[Private Environment.] When $\mathcal{D}$ is not available to the public in any way, the adversary may still able to exfiltrate data from the secure training environment by infecting the organization's software libraries that perform training. For example, in \cite{bagdasaryan2021blind}, the authors showed how an attacker can modify a training library to cause models that use it to learn a covert task in a black-box manner. For instance, they were able to modify the loss function in a library to cause a face counting model to covertly output the identity of the individual in an image if a special trigger (set of pixels) is placed within the input image.
\end{description}

\noindent The following are some example attack scenarios:
\begin{itemize}[itemsep=.2em]
    \item An attacker wants to steal bank transaction information from multiple financial institutions who are planning to use FL to collaborate on creating a powerful fraud detection model (e.g., \cite{UKUSSumm3:online,Federate79:online}). The attacker alters the distributed training code by compromising the member selected to be the orchestrator or by planting an insider in that member's organization. Alternatively, the attacker masquerades as a research member in the consortium and is selected as the orchestrator. Finally, the attacker receives the member's models and extracts the data from them.
    
    \item An attacker wants to steal a dataset in a DTaaS environment. The attacker enters the environment and trains an image classifier while covertly memorizing the dataset at the same time. The attacker then smuggles the dataset out with the model, since the exported model behaves legitimately on the expected classification task. 
    
    \item An attacker wants to steal private medical info from a DTaaS to blackmail people. The attacker trains a model to detect cancer in MRI scans that also memorizes the DICOM metadata of the patients stored in the dataset (used for labeling). The attacker then successfully exports the model because the model appears and functions as a cancer classifier. 

    \item An attacker wants to steal a company's latest datasets on a regular basis. The attacker performs a man-in-the-middle attack or uses an insider to make the company install tampered training libraries. The company's models are then unwittingly trained to memorize data which is then extracted by the attacker after deployment (e.g., from the company's products).
    
    \item An attacker wants to extract confidential information on specific individuals (e.g., fingerprints or the faces of a company's personnel). The attacker tampers with the installed libraries so that all of the models memorize their training data \cite{bagdasaryan2021blind}. The attacker then obtains a copy of the product with the embedded model (e.g., fingerprint sensor or camera) and executes the secondary task to extract the images.
\end{itemize}

In each of these cases, the attacker must keep the secondary task covert. This means that (1) the model work as usual (feed-forward execution) with good performance on the expected task, and (2) the model must present the expected architecture to avoid raising suspicion (e.g., the attacker cannot try to export an autoencoder when the expected task is classification).

Note that this data exfiltration attack is not an `inference-attack.' This is because the attacker is not querying the model to reveal information accidentally memorized/captured by the model. Rather, the attacker is using the model as a container to covertly exfiltrate knowledge/data out of a protected environment. In other words, the attacker extracts explicitly planted information by executing the secondary task and not by revealing unintentionally memorized samples by exploiting naturally occurring confidentiality vulnerabilities in the model.

We also note that although we focus on how a transposed model can be used to perform data exfiltration attacks via memorization, transposed models can be used perform other attacks as well (see Section \ref{appx:othersecondary} in the Appendix for examples).

%-------------------------------------------------------------------------------
\section{Transpose Attack}
%-------------------------------------------------------------------------------

In this section we formally define the transpose attack and describe how arbitrary deep neural networks can be transposed (trained and executed in reverse). We consider the secondary task of memorization as one possible secondary task of many. Therefore, in this section we discuss how transpose models work in general, and then in Section \ref{sec:mem} we discuss the secondary task of data memorization.

\subsection{Background}
Normally, a DNN is trained by optimizing the following objective function:
\vspace{-.5em}
\begin{equation}\vspace{-.5em}
       \arg \min_\theta \frac{1}{m}\sum_i^m \mathcal{L}\left( f_\theta(x_i), y_i\right)
\end{equation}
where $(x,y)\in \mathcal{D}_{train}$,  $|\mathcal{D}|=m$, and $\mathcal{L}$ is a differentiable loss function which measures error between the prediction $f(x)$ and the ground truth $y$.

The literature includes a number of studies that try to secretly learn a another function (e.g., \cite{guo2020hiding,bagdasaryan2021blind}). We refer to these attacks as \textit{hidden model attacks}. More formally, a hidden model attack is where a model $f_\theta$ is trained on a primary (expected) task, while another model $f'_{\theta^*}$ is trained on a secondary task where $\theta^* \subseteq \theta$, such that the execution of $f'_{\theta^*}$ is hidden from the defender. Both the primary and secondary tasks are embedded into $\theta$ by optimizing
\vspace{-.5em}
\begin{equation}\label{eq:objective}\vspace{-.5em}
       \arg \min_\theta \frac{1}{m}\sum_i^m \mathcal{L}^1\left( f_\theta(x_i), y_i\right) + 
       \lambda \frac{1}{m}\sum_i^m \mathcal{L}^2\left(  f'_{\theta^*} (x'_i), y'_i \right)
\end{equation}
where $\mathcal{L}^1$ and $\mathcal{L}^2$ are the loss functions used for the primary and secondary tasks respectively, $(x'_i, y'_i)$ is from dataset $\mathcal{D}$ which is required for learning the secondary task, and $\lambda$ provides a trade-off on the performance between the two tasks. In some cases,  $\mathcal{D} \subseteq \mathcal{D}_{train}$.

\begin{figure*}[t]
    \centering
    \includegraphics[width=\textwidth]{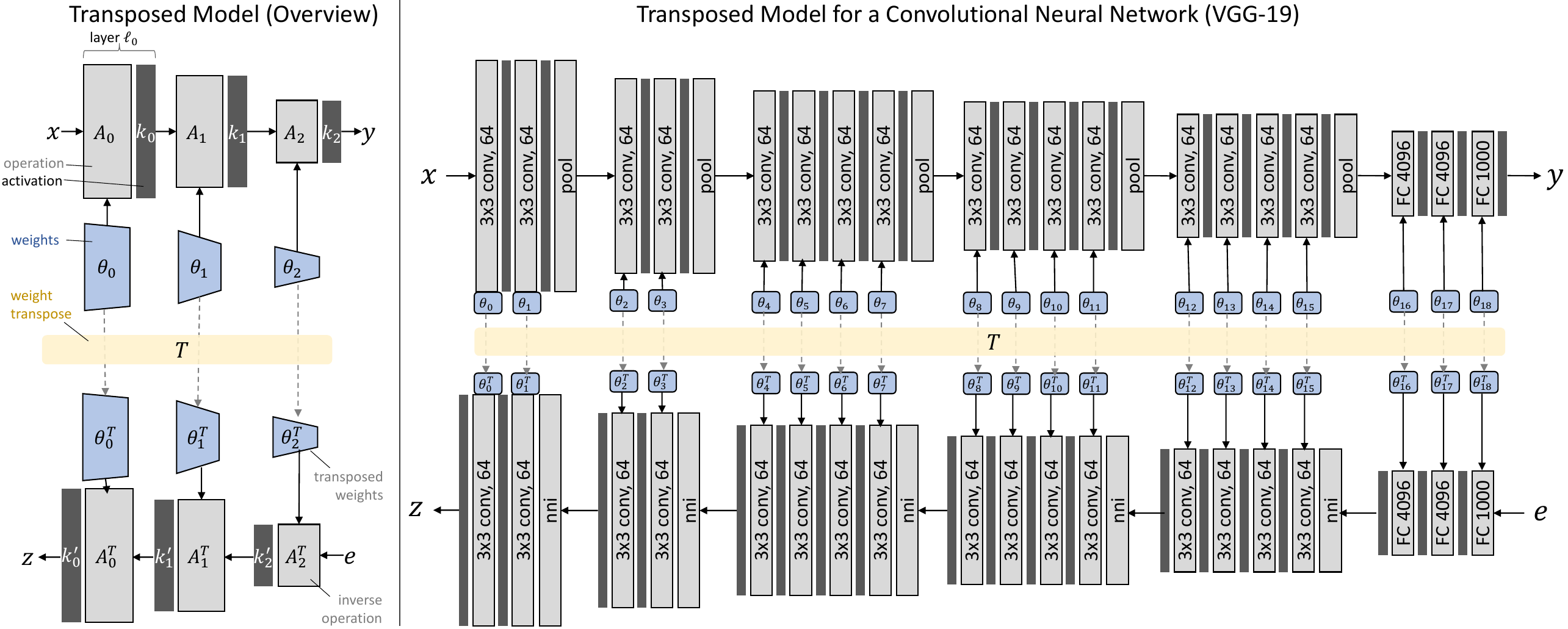}
    \vspace{-1.8em}
    \caption{Left - An overview of the transpose models: model $f_\theta$ is trained on the overt primary objective of $f(x)=y$, where $\theta= \{\theta_0,\theta_1,...\}$. In parallel, a transpose model $f'$ is trained on the secondary covert task of $f'_{\theta^T}(e)=z$, where $\theta^T= \{\theta_{m-1}^T,\theta_{m-2}^T,...\}$. The weights $\theta$ and $\theta^T$ are shared between the models during training. Therefore, the attacker can export the seemingly benign $\theta$ and then later recreate $\theta^T$ to use $f'$. Right: an example of how a CNN (VGG-19) is transposed.} 
    \label{fig:transpose}
    \vspace{-.5em}
\end{figure*}

\subsection{Backward Execution}
A transpose attack is a hidden model attack in which the primary task is performed in the forward direction as $f_\theta(x)$, while the secondary task is performed in the backward direction over $f$. To define how backward execution is achieved, we must first provide some notation on how a model is executed in the forward direction.

Let $\ell_i$ be a function which implements the $i$-th layer in a neural network such as a 2D convolution layer, pooling layer, or a fully connected (FC) layer. We denote $g_i$ as the output of $\ell_i$, while $g_0$ is the input to the network. Every layer has an associated set of weights $\theta_i$, an operation function $A_i$ (such as convolution), and an activation function $k_i$ (such as ReLU). We note that $A_i$ and $k_i$ can be the identity functions, meaning that the layer does not perform these actions.

In summary, the output of the $i$-th layer is:
\begin{equation}
g_i = \ell_i(g_{i-1}) = k_i(A_i(g_{i-1};\theta_i)).
\end{equation}
Note that $\ell_i$ has an input dimension of $dim(g_{i-1})$ and an output dimension of $dim(g_i)$.

To execute the secondary objective, we create a new model $f'$ which consists of the same layers but in the reverse order (see Fig.~\ref{fig:transpose}). Doing so has two challenges: (1) we must ensure consistent input-output dimensions between the layers, and (2) each layer must perform the `inverse' operation. To resolve these issues, we `transpose' each layer. To transpose layer $\ell_i$, we (1) obtain the transpose of $\theta_i$, denoted $\theta_{i}^T$, and (2) obtain the inverse operation of $A_i$, denoted $A_{i}^{-1}$. 

A forward pass with the transpose model $f'$ is expressed as
\begin{equation}
     g'_i = k'_i\left(A_{m-i}^{-1}(g'_{i-1};\theta_{m-i}^T)\right)  \hspace{2em} i=0,1,2,...,m.
\end{equation}
It is important to note that $\theta_i$ and $\theta_{i}^T$ are shared weights between $f$ and $f'$ where applicable. We also note that $k'$ can be any standard activation function (we typically use $k'=k$).
% K_{i}' is usually k_{i}

\subsection{Transposing a Layer}\label{subsec:layers}
The transpose of layer $\ell_i = k_{i}(A_i)$ is $k^{'}_i(A^{-1}_i)$. To invert an operation $A$, we must choose a suitable operation $A^{-1}$ such that the input and output dimensions are reversed ($A^{-1}:Y \rightarrow X$ where $A:X \rightarrow Y$). For parametric layers, this can be accomplished by transposing the parameters in $A$. 
In our study, we focus on operations found in common architectures: dense neural networks, convolutional neural networks, and transformer networks. 

For parametric operations, $A^{-1}$ must use the same weights as $A$. The following describes our implementation for these types of layers:
\begin{description}
    \item[Linear Layers (FC).] A linear layer (a.k.a. FC layer) performs the operation $A_{lin}(\mathbf{x};\theta_i)=\mathbf{x} \theta_i$, where $\mathbf{x} \in \mathbb{R}^N$ and $\theta_i \in \mathbb{R}^{N \times M}$. To transpose a linear layer, we only need to take the mathematical transpose of $\theta_i$. In other words, $\theta_{i}^T=\text{Tr}(\theta_i)$, and $A_{lin}^{-1} = A_{lin}$.

    \item[Convolution Layers.] These layers are typically used in vision models. For 2D convolution, the input is a tensor $\mathbf{x} \in \mathbb{R}^{C \times H \times W}$, where $H$ and $W$ are the spatial dimensions of $\mathbf{x}$, and $C$ is the number of channels.
    The operation $A_{conv}(\mathbf{x}; \theta_i)$ applies the bank of filters $\theta_i$ in strides over the spatial dimensions. The bank has the form $\theta_i \in \mathbb{R}^{M \times C \times K \times K}$, where $M$ is the number of filters, and $K$ is the size of the filter.
    
    To transpose a 2D convolutional layer, $A_{conv}^{-1}$ performs the deconvolution operation defined in \cite{zeiler2010deconvolutional}. To 'transpose' $\theta_i$, we permute the first two dimensions, such that $\theta_{i}^T \in \mathbb{R}^{C \times M \times K \times K}$. The same approach can be applied to convolution layers that have more or less spatial dimensions.
    
    \item[Transformer Blocks.] Blocks are a set of layers connected with a specific design. Transformer blocks (TBs) are scaled dot product attention units. They perform the operation $A_{trans}(\mathbf{q})$, where $\mathbf{q}$ is a sequence of tokens. Models that use TBs (a.k.a. transformer networks) can achieve state-of-the-art performance in natural language processing (e.g., BERT~\cite{devlin2018bert}) and vision tasks (e.g., Vision Transformer~\cite{dosovitskiy2020image, khan2021transformers}). Since the input and output dimensions of a TB are the same, there is no need to transpose the block.
\end{description}

For non-parametric operations, we chose operations which are often used to provide the reverse affect of $A$: 
\begin{description}
    \item[Pooling Layers.] CNNs often use pooling layers to reduce the dimensionality of signals propagated through the network. $A_{pool}(\mathbf{x})$ takes the average or maximum of non-overlapping patches in each channel $C$ of the input $\mathbf{x}\in \mathcal{R}^{C\times W \times H}$. As a result, the output has a reduced spatial dimensionality which is dependent on the patch sizes.     
    To transpose this layer, $A_{pool}^{-1}$ performs spatial up-sampling using nearest-neighbors interpolation~\cite{han2013comparison}.
\end{description}

\subsection{Transposing a Model}
Now that we know how to transpose a layer (Section \ref{subsec:layers}), we can transpose many different types of models. The general steps are to (1) reverse the order of the layers and (2) transpose each layer: replace each operation with its inverse operation and optionally replace the respective activation with an alternative activation. For example, if $f$ is a convolutional neural network such as VGG-19, then it consists of five blocks of convolution-pool layers followed by one block of FC layers. To obtain $f'$, all we need to do is reverse the order of the layers and obtain their transposed versions. The right side of Fig. \ref{fig:transpose} visualizes how the layers of a CNN (VGG-19) are transposed. This process results in a model $f(x)=y$ which can perform 1000-class image classification and a model $f'(e)=z$ which can perform a different task, such as image reconstruction. 

Similarly, consider a model $f$ which is a vision transformer (ViT) network used for image classification. In the forward direction, the model performs (1) image patch projection using a linear layer, (2) input marking with positional encoding, (3) mapping with a sequence of transformer blocks, and (4) prediction with pooling and FC layers. To obtain $f'$, we reverse the sequence of the layers mentioned above and transpose the FC and pooing layers. However, we also move the positional encoding layer to the front of the sequence of transformer blocks.

\subsection{Model Training}
To train a transpose model $f$, we train $f_{\theta}$ and $f'_{\theta^T}$ in tandem over the shared weights $\theta$. During training, each model is optimized according to its own objective (see equation \ref{eq:objective}). For example, if $f$ is a classifier, and $f'$ is a memorization model, then $f$ may use cross-entropy loss, and $f'$ may use $L_2$ loss.

The complete training process is presented in Algorithm~\ref{alg:training}. In lines \ref{line:trans1} and \ref{line:trans2} we transpose the model back and forth to alternate between $f_{\theta}$ and $f'_{\theta^T}$. This is done for clarity but would not be done in practice. This is because the transpose operation on $\theta$ is mutable.

\begin{algorithm}
	\caption{Transpose Model Training} \label{alg:training}
	\begin{algorithmic}[1]
		\For {$epoch=1,2,\ldots$}
			\For {$(X, Y) \in \mathcal{D}_{train} $} \Comment{draw batch}
				\State $Y_{pred} \leftarrow f_{\theta}(X)$
                \State loss1 $\leftarrow \mathcal{L}^1(Y,Y_{pred})$
				\State $\theta \leftarrow$ optimize($\theta$, loss1) \Comment{iteration of GD}
                \State $(X', Y') \leftarrow$ drawNextBatch($\mathcal{D}$) \Comment{draw batch}

                \State $f'_{\theta^T} \leftarrow $transposeModel$(f_{\theta})$ \label{line:trans1}
                \State $Y'_{pred} \leftarrow f'_{\theta^T}(X)$
                \State loss2 $\leftarrow \mathcal{L}^2(Y',Y'_{pred})$
				\State $\theta^T \leftarrow$ optimize($\theta^T$, loss2) \Comment{iteration of GD} 
                \State $f_{\theta} \leftarrow $transposeModel$(f'_{\theta^T})$
                \label{line:trans2}
			\EndFor
		\EndFor
	\end{algorithmic} 
	
\end{algorithm}

%-------------------------------------------------------------------------------
\section{Data Memorization}\label{sec:mem}
%-------------------------------------------------------------------------------
In this section we propose a novel method for teaching a neural network to memorize samples so that they can be \textit{systematically} retrieved. This is an example of a secondary task that can be used in a transpose attack.

It is well known that DNNs can be intentionally taught to memorize samples. For example, autoencoders are neural networks that are designed to reconstruct samples from encodings~\cite{theis2017lossy}. However, models like autoencoders learn implicit codes that cannot be easily determined (found) after training. To address this limitation, we propose a method for teaching a model to become a data retrieval system. 

The objective of data memorization is to approximate the function $h(e_i)=x_i$, such that $e_i$ is an index that points to sample $x_i \in \mathcal{D}$. In this task, the index $e_i$ should be deterministic so that it can be used to iterate over all items in $\mathcal{D}$. The model $h_\theta$ is fitted using conventional machine learning tools. However, in contrast to conventional machine learning, $h$ does not have the objective of generalizing to unseen samples. In other words, the objective is to intentionally overfit to the dataset $\mathcal{D}$.

We will now describe how we design the indexer and train the model $h_\theta$.

\subsection{Spatial Indexing}
In order to index items stored in $h_{\theta}$, we propose using a spatial index. Let $I: \mathbb{N}_0 \rightarrow \mathbb{R}^n$ be a function which maps a natural number (the index value) to a point in an $n$-dimensional euclidean space, where $I(i) \neq I(j)$ $\forall i,j$, where $i\neq j$. We refer to this function as an indexer and its outputs as spatial indices. With an indexer, a user can systematically find every indexed point in $\mathbb{R}^n$ by executing the sequence $I(0)$, $I(1)$, $I(2)$, etc... 

One implementation of $I$ is to use binary enumeration. For example, with a range of $\mathbb{R}^3$, we would obtain $I(0)=000$, $I(1)=001$, $I(2)=010$, and so on. Although this is convenient, it would restrict $I$ to $2^n$ spatial indices. To increase the domain of $I$, we can use $n$-ary values (e.g., decimal, hex., etc.).

Gray code is an ordering of the binary numeral system such that neighboring numbers have only a difference of one bit between them. We found that Gray code increases the memorization capacity compared to using binary. This is because Gray code produces a mapping that is dense, which helps $h_\theta$ better compress information. For example, consider the euclidean distance between the values $15$ and $16$ in binary ($01111$ and $10000$) and Gray code ($01000$ and $11000$). In our work, we use $n$-ary Gray code to increase the the domain of $I$ in a dense manner. Therefore, to produce a spatial index for item $x_i$, we perform
\vspace{-.5em}
\begin{equation}
    I(i) = Gray(i) 
\end{equation}
where $i \in \mathbb{N}_0$ is the index value.

To increase capacity further, we borrow from the concept of embeddings. An embedding is a vector $v_a \in \mathbb{R}^n$, where $v_a$ represents the object $a$. The value of $v_a$ is chosen such that if $a$ is similar to another object $b$, then $\Vert v_a - v_b \Vert_2$ will be small and vice versa. Neural networks work well with embeddings, because the model can internally use the fact that the euclidean distance captures similarity. Guided by this intuition, we found that the memorization capacity can be increased if the spatial index of similar items are near each other. This enables the model to compress similar patterns using fewer weights.

\begin{figure}
    \centering
    \includegraphics[width=\columnwidth]{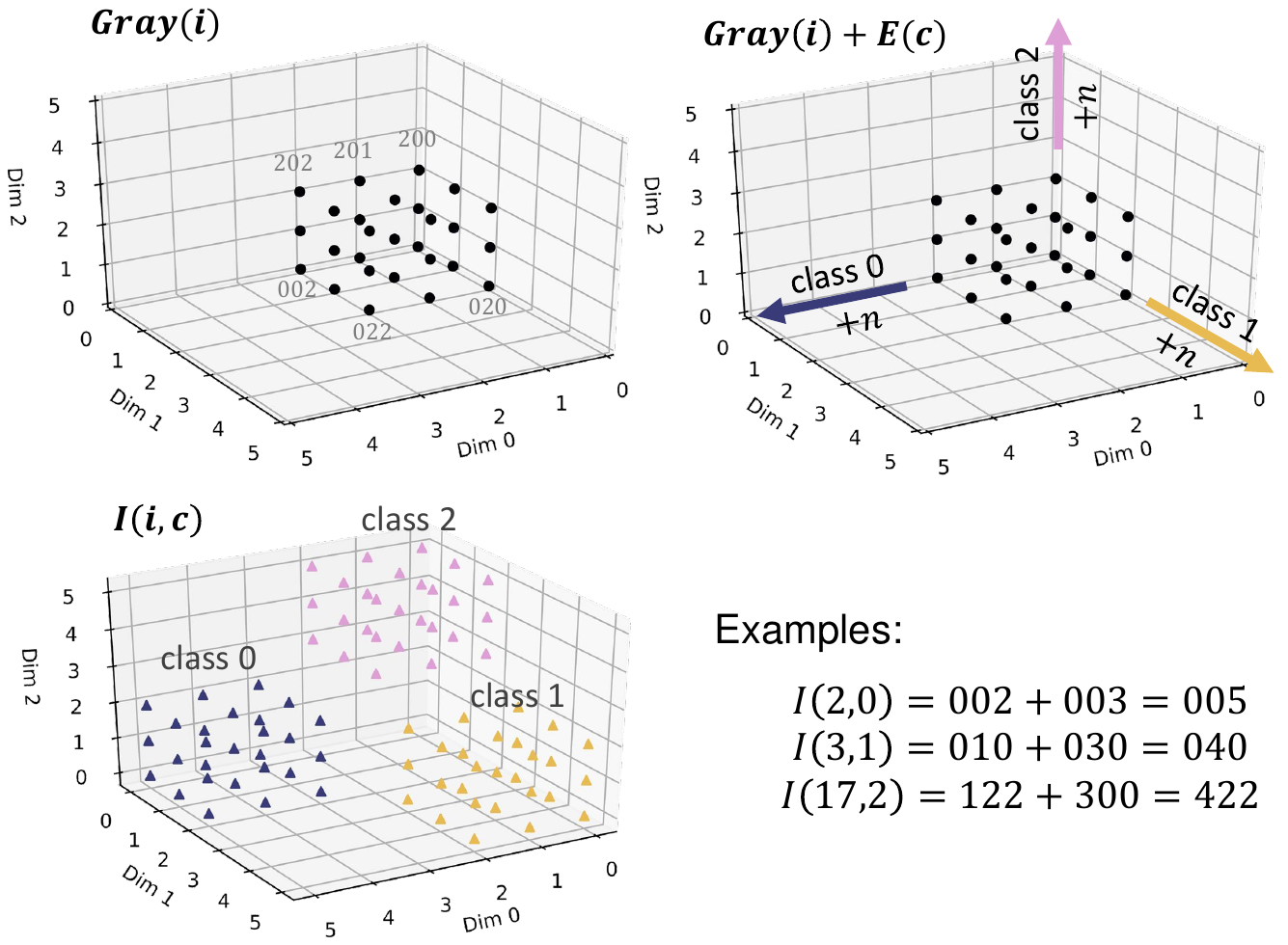}
    \vspace{-2em}
    \caption{An example of $n$-dimensional spatial index $I$ where $n=3$. Given the $i$-th item in class $c$, the spatial index is calculated by (1) finding the $n$-ary Gray code for $i$ and then (2) adding to it the one-hot encoding for $c$ (multiplied by $n$). In this example, only $n^n=27$ items can be indexed per class.}
    \label{fig:Sample_encoding}
    \vspace{-1em}
\end{figure}

We now present how this can be used to improve indexing. Let $C$ be the set of all classes in $\mathcal{D}$ (or some other attribute that clusters items). Let $E(c)$ be an embedding function which maps each $c \in C$ to a unique vector $v_c \in \mathbb{R}^n$. This vector can be used to project (offset) the spatial indices of each class to their own regions. Finally, the complete indexer is defined as:
\begin{equation}\label{eq:indexer}
    I(i,c) = Gray(i) + E(c)
\end{equation}
Note that $I(i,c)$ is the spatial index to $i$-th item in class $c$. In this work, we implement $E$ in two different ways. One way is to use one-hot encodings for each class multiplied by $n$, where $n$ is the value used in the $n$-ary Gray code. This ensures that indices between classes are orthogonal and do not intersect. Fig.~\ref{fig:Sample_encoding} visualizes how our spatial indexer works with the one-hot encodings scheme when $n=3$ and $|C|=3$. The second way is to use random embeddings, where each embedding is mapped to a specific class. 
The advantage of using random embeddings is that the number of classes does not restrict input size of $h$.

\subsection{Memorization Training Objective}
Let $x_{ic}$ be the $i$-th sample from class $c$. The memorization model $h_\theta$ can be seen as a generator which generates $x_{ic}$ from index $I(i,c)=e_{ic}$. Therefore, we train $h_\theta$ by solving the following optimization problem:
\begin{equation}
    \arg \min_\theta  \sum_{c} \sum_{i} ||h_\theta(I(i,c)) - x_{ic}||_2
\end{equation}
In other words, we train $h_\theta$ like a regular DNN using $L_2$ loss between the generated sample and the expected sample for the given spatial index.

%-------------------------------------------------------------------------------
\section{Evaluation}
%-------------------------------------------------------------------------------
In this section we evaluate transpose attacks. Specifically, we focus on evaluating the secondary task of data memorization where the target $\mathcal{D}$ is a dataset of images and $\mathcal{\tilde{D}}$ are the retrieved images (reconstructed by $f'$). For reproducibility, readers can download our source code for the transposed model memorization attack.\footnote{\url{https://github.com/guyAmit/Transpose-Attack-paper-NDSS24-/tree/main}}

In a data exfiltration attack, the attacker wants to either (1) breach the data's confidentiality, or (2) steal intellectual property (IP). To evaluate the attack's ability to breach confidentiality, we analyze the quality of the retrieved images at various granularity levels. To assess the stolen data's utility, we train new model on the retrieved data and measure its performance. These experiments are covered in Sections \ref{subsec:conf} and \ref{subsec:ip}. In these sections we also discuss how the number of memorized images impacts the performance on the primary and secondary tasks.

It is reasonable to assume that models with more weights will be able to memorize more. In Section \ref{subsec:capacity} we investigate which model hyperparameters (for example, the number of layers or the size of the layers) contribute to increased capacity. 

Finally, in Section \ref{sec:mem} we suggested a spatial index consisting of multiple components. To demonstrate the contribution of each of these components, we perform an ablation study, which is discussed in Section \ref{subsec:ablation}.

\subsection{Experiment Setup}
\label{sec:exp_setup}
\begin{description}[itemsep=.2em]\setlength{\parskip}{0pt}
    \item[The Attack.] We explore a transpose attack on model $f_\theta$ where the primary task $f$ is image classification on $\mathcal{D}_{train}$, and the secondary task $f'$ is data memorization of $\mathcal{D} \subseteq \mathcal{D}_{train}$. We explore this scenario with various different configurations and settings. In all cases, $f$ is trained on a static number of samples (all of $\mathcal{D}_{train}$); however, the size of $\mathcal{D}$ varies depending on the experiment.

    \item[Datasets.] We used a variety of different image datasets in our evaluations: MNIST~\cite{deng2012mnist}, CIFAR-10~\cite{krizhevsky2009learning}, and CelebA~\cite{liu2015faceattributes}. MNIST is a classic handwritten digit classification dataset. CIFAR-10 is an image classification dataset consisting of 10 different classes. Finally, CelebA is a dataset of face images where each image is annotated with both attributes (has glasses, is smiling, ...) and an identity. 

    \item[Architectures.] We evaluated transpose attacks on three very different architectures: fully connected (FC) networks, convolutional neural networks (CNN) and vision transformer networks (ViT). We examined various different configurations for each of these architectures. Additional details about the architectures are provided below. 
    
    \item[Training.] Training was performed using Algorithm \ref{alg:training}. Models trained on the CIFAR-10 and MNIST datasets were given 500 training epochs for the primary task. We implemented early stopping in the secondary task if the $L_2$ loss stopped improving. Both $f$ and $f'$ used batch sizes of 64. For models trained on CelebA, we fine-tuned the backbone of a facial recognition model. This was done to reduce the training time. During the attack, the model was then fine-tuned for both tasks over 40 epochs. Here, batch sizes of 32 were used due to GPU memory limitations.

    \item[Metrics.] For the primary task, accuracy (ACC) was used to measure the performance, and for the secondary task, two other measures were used. The first is the mean squared error (MSE) which we take between the retrieved image and the original: $\frac{1}{n}\sum_{i}^{n}(f'_{\theta^T}(I(i,c)) - x_i)^2$. A low MSE value indicates that the retrieved image's pixels are accurate and that the image has high-fidelity. We also refer to this metric as ``pixel accuracy.'' 

    Sometimes, an image may not be accurate pixel-wise but still contain confidential information. For example, if an individual's face is retrieved by $f'$ from CelebA, but the face is off-center, the MSE will be low, but the confidentiality has still been breached. Therefore, we also measure the structure similarity (SSIM)~\cite{7025115} and ``feature accuracy'' in our experiments. Feature accuracy is the performance of a highly accurate model trained for $\mathcal{D}_{train}$ classification.
    For MNIST we used an FC model that has 99\% accuracy, for CIFAR-10 we used a Resnet18 model~\cite{he2016deep} with 95\% accuracy, and for CelebA we used a ViT model that obtains a DICE score of 74\% (DICE is equivalent to accuracy in multi-label classification).

    %excluding the samples memorized by $f'$. 
    
\end{description}

\begin{figure*}[t!]
\setlength\tabcolsep{1pt}
\settowidth\rotheadsize{CelebA ViT}
\setkeys{Gin}{width=\textwidth}
\begin{tabularx}{0.9\linewidth}{l c }% <-- here is determined table width

\rothead{\texttt{MNIST-FC}} &  \includegraphics[valign=m, width=0.96\linewidth]{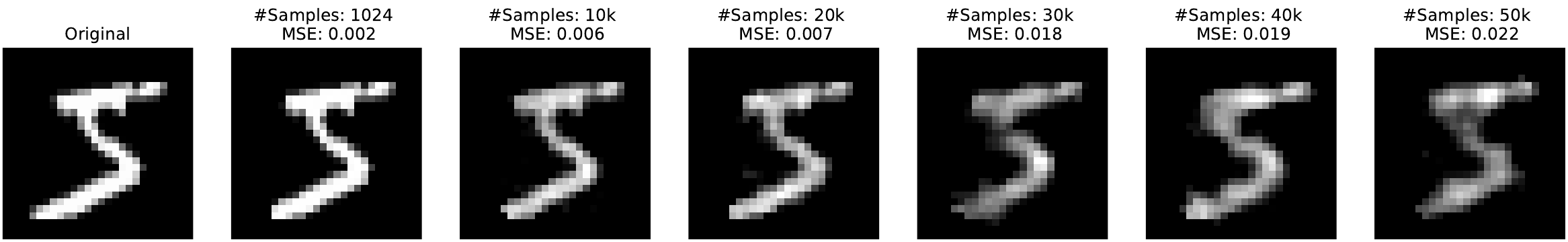}
\\  
    \addlinespace[2pt]
\rothead{\texttt{MNIST-CNN}} &  \includegraphics[valign=m, width=0.96\linewidth]{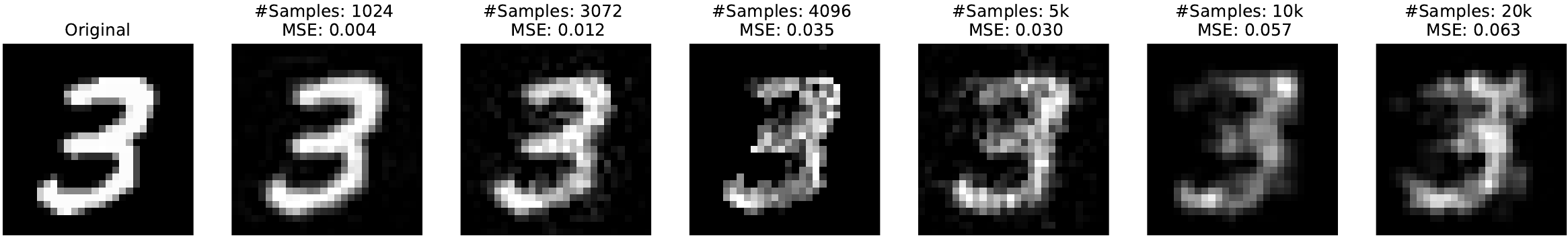}
\\  
    \addlinespace[2pt]

\rothead{\centering
         \texttt{CIFAR-CNN}}        &   \includegraphics[valign=m, width=0.96\linewidth]{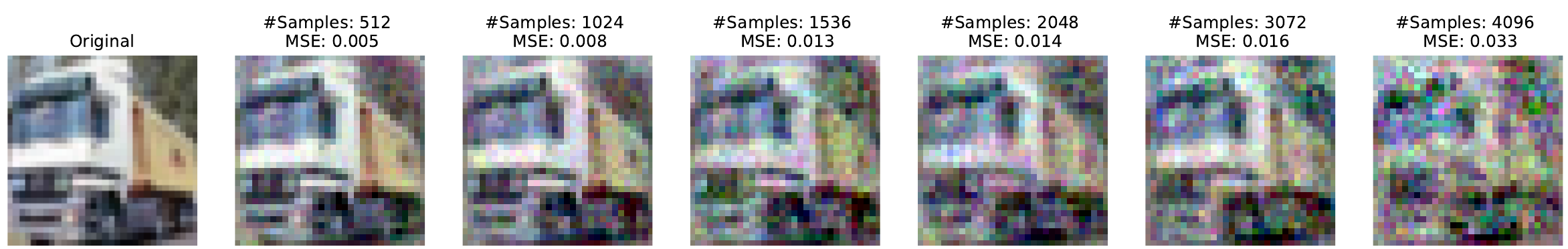}
                         \\  
    \addlinespace[2pt]
\rothead{\texttt{CIFAR-ViT}} &  \includegraphics[valign=m, width=0.96\linewidth]{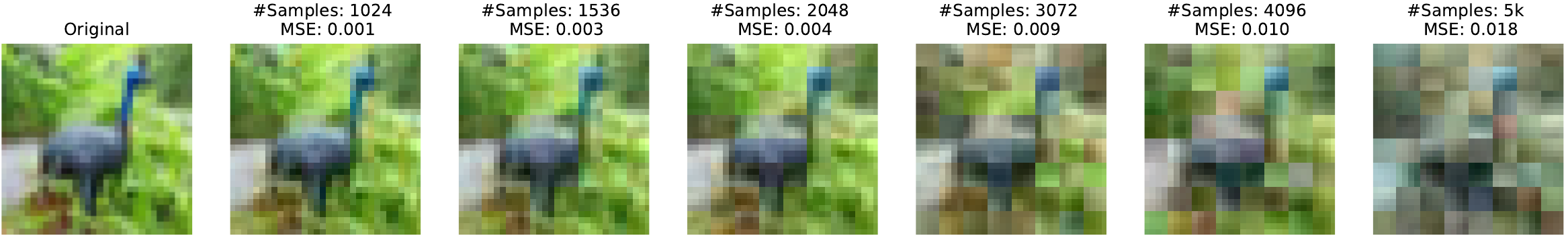}
\\

    \addlinespace[2pt]
\rothead{\texttt{CelebA-ViT}} &  \includegraphics[valign=m, width=0.96\linewidth]{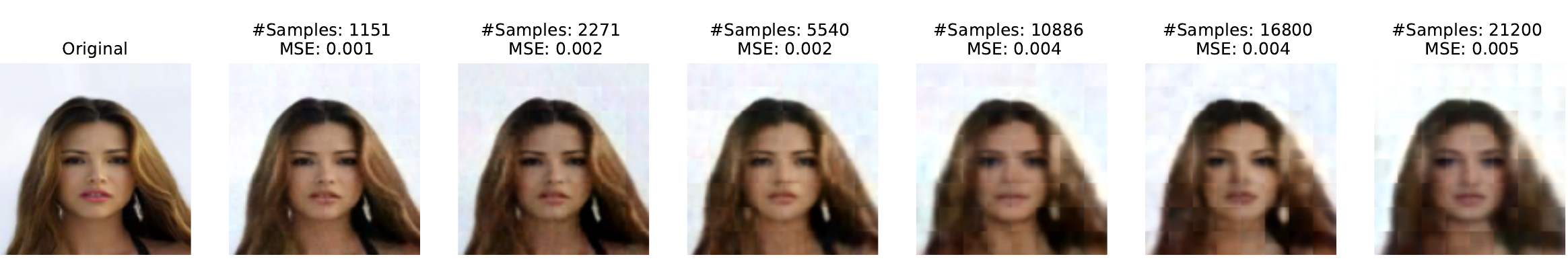}
\end{tabularx}

\caption{Samples of images retrieved using $f_{\theta^T}$ with different models. For example, \textit{``give me the 472nd image for class car''} or \textit{``give me the 15th image of identity A.''} Left to right: When we increase the number of images that $f_{\theta^T}$ must memorize, the quality of the retrieved images degrades.}
\label{figure:quality_change}
\vspace{-1em}
\end{figure*}

\begin{figure*}[t]
\centering
\setlength\tabcolsep{3pt}
\settowidth\rotheadsize{CelebA}
\setkeys{Gin}{width=\hsize}
\begin{tabularx}{1\linewidth}{l XXXX }
\rothead{\centering
         MNIST}        &   \includegraphics[valign=m]{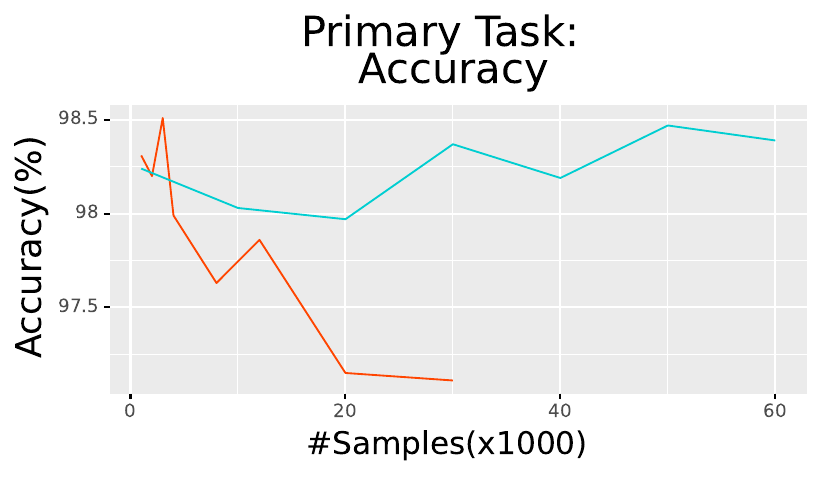}
                        &   \includegraphics[valign=m]{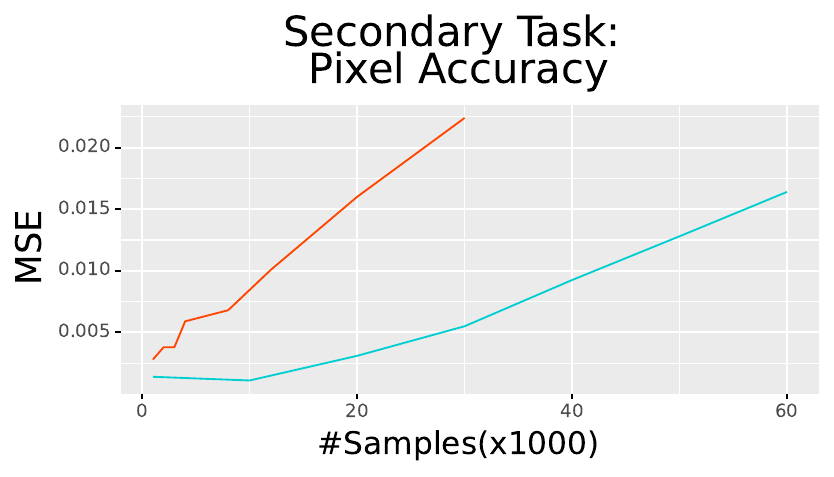}  
                        &  
                        \includegraphics[valign=m]{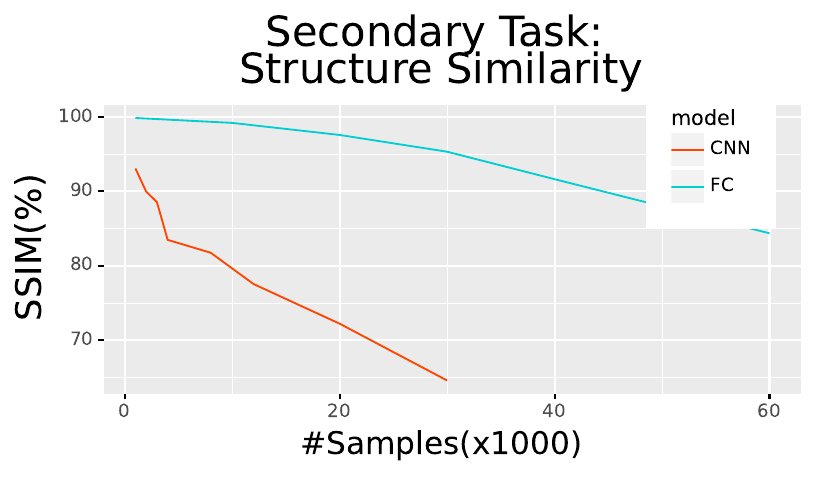}  
                        &  
                        \includegraphics[valign=m]{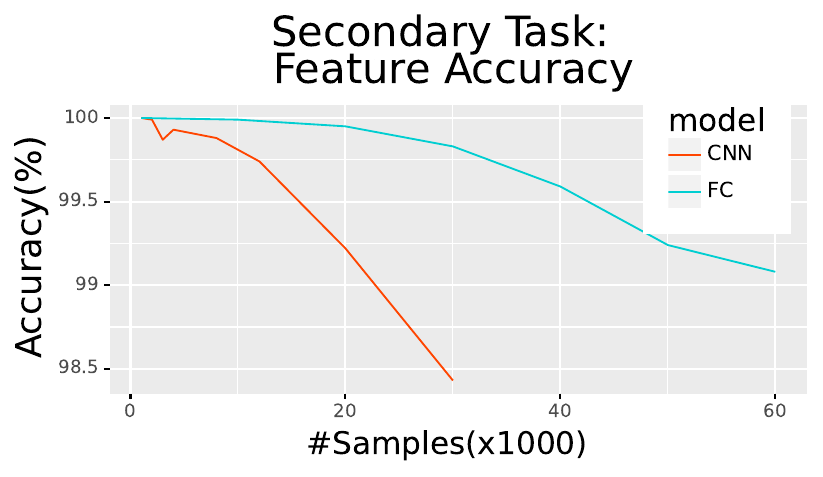}     \\  
    % \addlinespace[2pt]
    
\rothead{\centering CIFAR}       &      \includegraphics[valign=m]{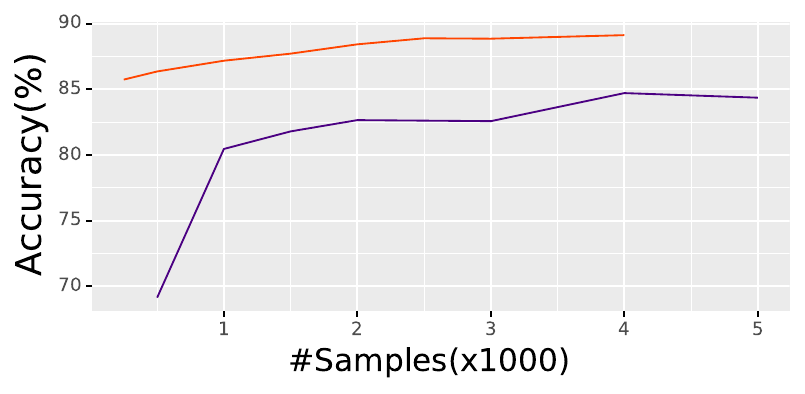}
                        &   \includegraphics[valign=m,  ]{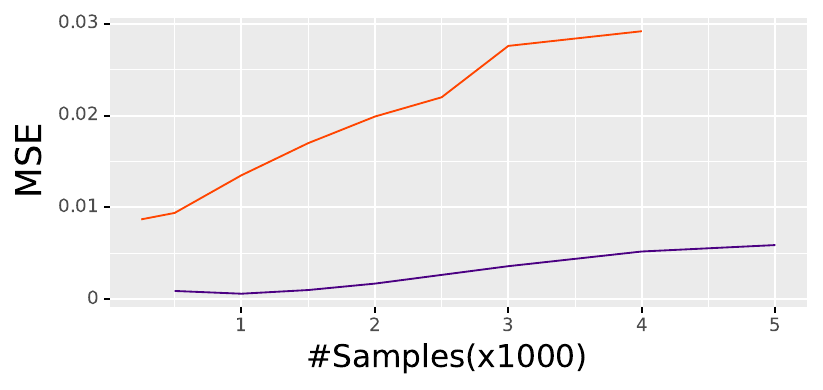}
                        &
                        \includegraphics[valign=m]{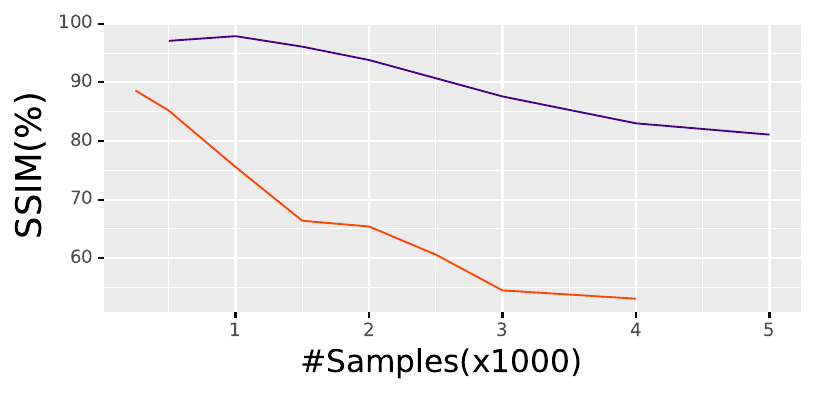}  
                        &                         
                        \includegraphics[valign=m,  ]{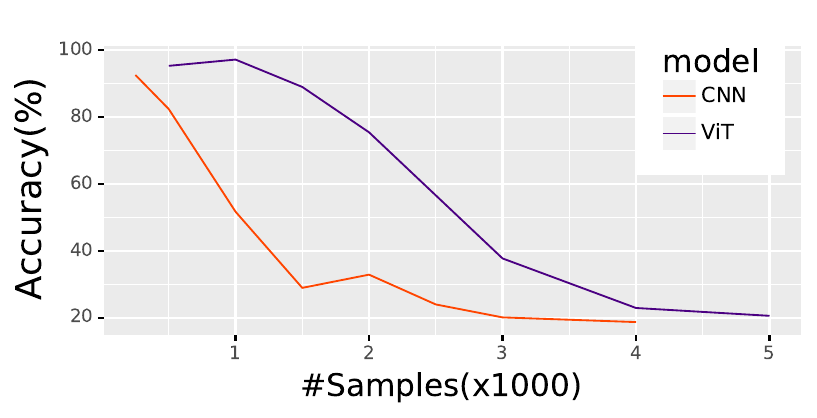} \\  
    \addlinespace[2pt]
\rothead{\centering CelebA}       &      \includegraphics[valign=m,  ]{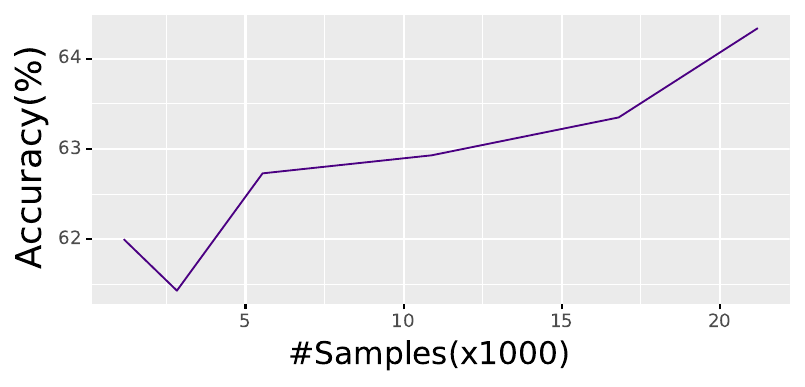}
                        &   \includegraphics[valign=m,  ]{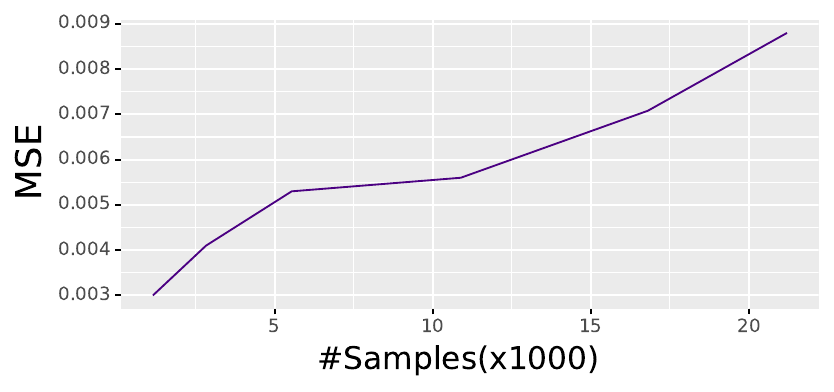}
                        &   
                        \includegraphics[valign=m]{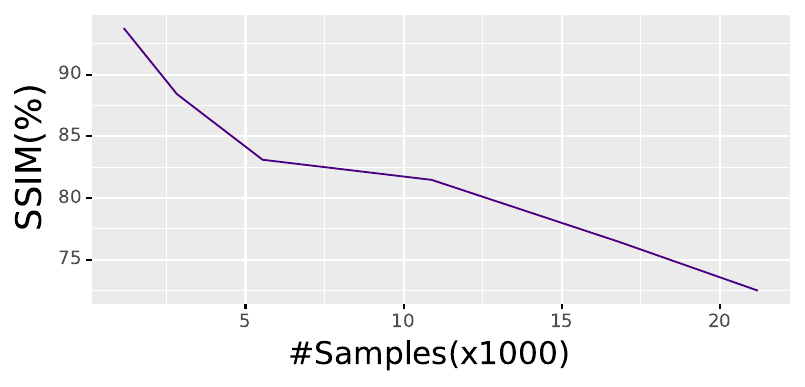}  
                        & 
                        \includegraphics[valign=m,  ]{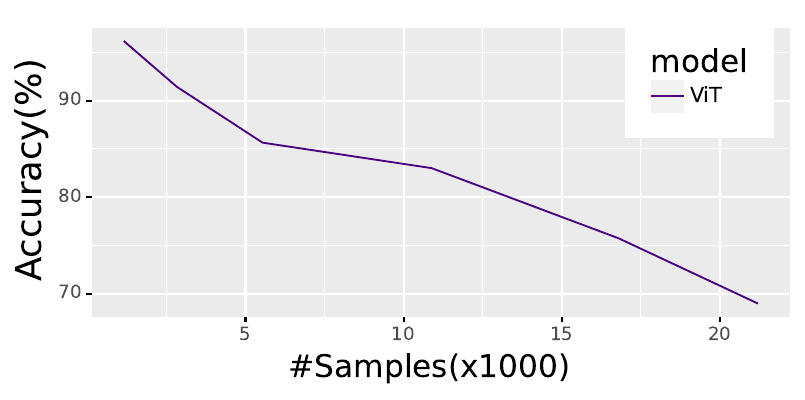}
\end{tabularx}
\caption{The trade-off between the performance on the primary and secondary tasks as a function of the number of samples memorized by $f'$. Rows: Models grouped by their datasets. Columns: The performance on the primary and secondary tasks. Pixel accuracy is MSE and feature accuracy is the accuracy of different classifier (described in~\ref{sec:exp_setup}) executed on the retrieved images. Each data point is the average result from five random runs.}
\label{fig:Confidentiality}
\vspace{-1em}
\end{figure*}

\subsection{Image Quality (Confidentiality)}\label{subsec:conf}
\textbf{Setup.} 
To assess the quality of images retrieved from $f'_{\theta^T}$, we explored both subjective (visual) quality and the objective (measured) quality as a trade-off between performance and the number of samples memorized. Several different models were used in this experiment:

\textbf{Models.} \texttt{MNIST-FC} is a fully connected network with three layers, each of which has 1024 neurons, trained on MNIST. 
\texttt{MNIST-CNN} is a convolutional neural network with three layers, each of which has 128 channels, trained on MNIST.
\texttt{CIFAR-CNN} is a convolutional neural network with three layers, each of which has 384 channels, trained on CIFAR-10.
\texttt{CIFAR-ViT} is a transformer network with seven layers, a patch embedding size of 384, and an MLP dimension of 3x384 with 12 heads, which is trained on CIFAR-10.
\texttt{CelebA-ViT} is a transformer network with 20 layers, a patch embedding size of 512, and an MLP dimension of 2048 with eight heads, which is trained on CelebA.

We intentionally chose medium sized models due to time constraints, since each model had to be retrained a number of times in our experiments. An evaluation of larger models is presented in Section \ref{subsec:capacity}.

\textbf{Attack Implementation.}
Each of the transposed models used the respective dataset's classes in the spatial indexer for $E(c)$ (equation \ref{eq:indexer}). The exception is \texttt{CelebA-ViT} where $f$ was trained to classify properties of the face (wearing hat, eye glasses, ...), and $f'$ used the identities of the faces instead. In other words, $f$ appeared as a face attribute classifer, but $f'$ could be used to retrieve the $i$-th face of identity `A.'
Regarding the implementation of $E(c)$: models trained on MNIST and CIFAR-10 used the one-hot embedding method. The model trained on CelebA used the random embedding method. The reason for this is that the output of \texttt{CelebA-ViT} has a size 40. However, the input to the model's transposed version of this model need to be larger than 40 because there are more than 40 classes (identities). The results of an ablation study comparing the two projection methods are presented in Section \ref{subsec:ablation}.

\textbf{Evaluation Approach.} 
For the visual evaluation, we trained a new model for each target amount of memorized images. For the trade-off evaluation, five models were trained each time on random target images and the results were averaged.

\textbf{Results.}
We found that as we increase the target number of memorized images, the quality of the retrieved images starts to degrade. Fig. \ref{figure:quality_change} demonstrates this observation using samples of retrieved images, where \#\textit{samples} is the number of samples memorized by $f'$ (i.e., $|\mathcal{D}|$). The figure shows that even with rather small architectures, we were able to retrieve a large number of high-quality images. For example, the \texttt{MNIST-FC} model was able to memorize the \textit{entire} training set of 60K samples. The \texttt{MNIST-CNN} was able to memorize at least 33\% of the data, which is understandable, since it has fewer parameters than an FC network. The CIFAR-10 dataset is far more complex with many details in the background. This made it harder for the CNN and ViT models to find common patterns to compress and store in $\theta^T$. Regardless, they were still able to store at least 5,000 images and retrieve them with recognizable content. We also observed that a `patching' artifact appeared in the images retreived from \texttt{CIFAR-ViT} at a certain point. This is due to the process ViT architectures use to project regions of inputs. With \texttt{CelebA-ViT}, is able to memorize and retrieve at least 21200 samples with good quality. In Table \ref{tab:CelebA}, the number of identities memorized by each \texttt{CelebA-ViT} model presented in Fig. \ref{figure:quality_change} is listed.

%tab:CelebA
\begin{table}[t]
\caption{The number of identities memorized from CelebA. We memorize all samples per target identity.}
\vspace{-1em}
\label{tab:CelebA}\vspace{1em}
\centering
\begin{tabular}{ccccccc}\hline\hline
\#Samples    & 1151 & 2271 & 3380 & 4460 \\ 
\#Identities  & 40   & 80   & 120  & 160 \\ \hline

\#Samples    & 5540 & 10886 & 16K & 21K\\ 
\#Identities  & 200   & 400   & 600  & 800\\ \hline\hline

\end{tabular}
\vspace{-1em}
\end{table}

The trade-off between the performance on the primary and secondary tasks (as a function of the number of images memorized) is presented in Fig. \ref{fig:Confidentiality}. The first column measures the primary task performance in terms of classification accuracy and the last three columns evaluate the secondary task in terms of pixel accuracy (MSE), structure similarity (SSIM) and feature accuracy (via an auxilary model). The general trend is that the image quality degrades as more images are added.\footnote{Please use Fig. \ref{figure:quality_change} as a reference for each model's MSE value.}
% Not better than baseline model
% Early stopping
%Interestingly, memorizing more samples generally improves the performance of the primary task. We leave the investigation of positive uses for transpose models to future work. 
Finally, as the number of memorized images increases, the structure similarity remains relatively high but the feature accuracy drops significantly. This is because the content is still recognizable but the key features which the auxiliary model relied on were lost. For example, ViT models tend to form a patching artifact when memorizing many images but the content is still quite interpretable. Overall, Fig. \ref{fig:Confidentiality} indicates that an adversary would most likely be concerned with maximizing the number of samples to memorize and care less about the performance of $f$ to evade detection during export control.% (see Fig. \ref{fig:attack_model}).

We note that Fig. \ref{fig:Confidentiality} seems to indicate that increasing the number of memorized samples increases the primary task performance. However, this is only because the number of training iterations is increases as more samples are memorized; a side-effect of using early-stopping on the secondary task's performance.

In summary, there is a trade-off between image quality and the number of memorized images. However, it appears that if the images in $\mathcal{D}$ have many shared features then the model can compress more samples in the same weights (e.g., \texttt{CIFAR-ViT} vs \texttt{CelebA-ViT}).

\subsection{Data Reuse (IP Theft)}\label{subsec:ip}

\begin{table*}[t]
\caption{The utility of the stolen images when used to train new models. The new models were an FC for MNIST (left), a ResNet18 for CIFAR-10 (center), and a ViT for CelebA (right). Here, the `*' in $\tilde{\mathcal{D}}_{*}$ is the transposed model used to steal the training data. 
}
\label{table:IP_theft}\vspace{1em}
\centering

\begin{tabular}{ccc}

\begin{tabular}{cccc}
\multicolumn{4}{c}{MNIST-FC}                                                                              \\ \hline \hline
\multirow{2}{*}{\# samples} & \multicolumn{3}{c}{Accuracy when trained on:}                            \\
                            & $\mathcal{D}$ & $\tilde{\mathcal{D}}_{FC}$ & $\tilde{\mathcal{D}}_{CNN}$ \\ \hline
2048                        & 92.04         & 92.09                      & 91.95                       \\
10K                         & 96.99         & 96.91                      & 93.94                       \\
20K                         & 98.07         & 97.95                      & 92.21                       \\
30K                         & 98.44         & 98.19                      & 85.96                    \\  \hline\hline
\end{tabular}
&
\begin{tabular}{cccc}
\multicolumn{4}{c}{CIFAR-ResNet18}                                                                           \\ \hline  \hline 
\multirow{2}{*}{\# samples} & \multicolumn{3}{c}{Accuracy when trained on:}                             \\
                           & $\mathcal{D}$ & $\tilde{\mathcal{D}}_{CNN}$ & $\tilde{\mathcal{D}}_{ViT}$ \\ \hline 
1024                       & 51.75         & 46.63                       & 52.84                       \\
2048                       & 66.44         & 34.02                       & 63.85                       \\
3072                       & 76.6          & -                           & 61.59                       \\
4096                       & 78.53         & -                           & 61.19                      \\  \hline  \hline 
\end{tabular}%
&
% \centering
% \vspace{1em}
\begin{tabular}{cccc}
\multicolumn{3}{c}{CelebA-ViT}                                                                           \\ \hline  \hline 
\multirow{2}{*}{\# samples} & \multicolumn{2}{c}{Accuracy when trained on:}                             \\
                           & $\mathcal{D}$ & $\tilde{\mathcal{D}}_{ViT}$ \\ \hline 
5K                       & 60.35         & 60.55  \\
10K                       & 63.58         & 62.33 \\
16K                       & 65.87          & 63.23 \\
21K                       & 65.63         & 64.33   \\  \hline  \hline 
\end{tabular}%
\end{tabular}

\end{table*}
\textbf{Setup.} In addition to examining confidentiality breaches resulting from this attack, it is important to understand the utility of the images stolen using the attack. We want to demonstrate that an attacker can train new models on the reconstructed dataset. To measure this, we train a new model on $\tilde{\mathcal{D}}$ and compare its performance to benign models trained on the original samples $\mathcal{D}$. 

In this experiment we explore two cases: stealing MNIST and stealing CIFAR-10 to train a new model. The models used in the previous experiment (see Section \ref{subsec:conf})) were also used here to perform this theft. Regarding the new models, we used a three-layer FC network on the reconstructed MNIST datasets and a ResNet18~\cite{he2016deep} on the reconstructed CIFAR-10 datasets. We then explored the impact of $|\mathcal{D}|$ on the performance of these models.

\textbf{Results.} Table~\ref{table:IP_theft} presents the results of this experiment. The results indicate that transpose memorization attacks can provide utility to the adversary. The margin between the baseline model (trained on $\mathcal{D}$) and the adversary's model (trained on $\tilde{\mathcal{D}}$) varies between 0.05\% and 12.5\%. In general, the margin increases with the the size of $\mathcal{D}$.
This presents a challenge to the adversary: models perform better when trained on more data, however memorizing more data harms the quality of the training data $\tilde{\mathcal{D}}$. As a result, is is preferable to obtain fewer high-quality samples than many low-quality samples. 

If the adversary can perform multiple model exports, then they can focus on quality over quantity per export and extract \textit{all} of $\mathcal{D}_{train}$'s samples at a high-quality. MNIST, CIFAR-10 and CelebA have 60K, 50K and 160K training samples respectively. When the ideal transpose models are chosen from Table \ref{table:IP_theft}, it takes only 2 exports with \texttt{MNIST-FC} and 6 exports with \texttt{MNIST-CNN} to extract all of MNIST. To extract all of CIFAR-10, it takes 25 exports with \texttt{CIFAR-ViT} and 50 exports with \texttt{CIFAR-CNN}. Similarly with \texttt{CelebA-ViT}, it takes 10 exports to extract all of CelebA. 

In summary, from this experiment we learn that an adversary who seeks to use this attack to breach confidentiality may try to exfiltrate many low-quality samples; in contrast, an adversary trying to gain utility will try to exfiltrate fewer high-quality samples.

\subsection{Model Size \& Memorization Capacity}\label{subsec:capacity}

\textbf{Setup.} Intuitively, the more weights a model has, the more data it can memorize. This is because $f'$ learns to compress $\mathcal{D}$ into $\theta$ in the form of feature maps. Therefore, having more parameters should increase the memorization capacity. However, an adversary cannot just export an unreasonably large model, since this may raise suspicion. A defender could also put an export limit on the model size. In this experiment, we investigate which aspects of a model contribute towards increasing the capacity of transpose model performing memorization.

We explore two dimensions: model depth (number of layers) and model width (number of neurons per layer). Our hypothesis is that more layers contribute to improved compression of common features, while more neurons per layer contribute to the model's ability to memorize a greater variety of images. We explore this concept using the MNIST and CIFAR-10 datasets. We also consider the hypothesis that the depth and width are not correlated to capacity, rather that the memorization capacity is solely dependent on the total number of weights. %This experiment is performed over all of our models except CelebA due to time limitations.

\begin{figure}[t]
    \centering
    \includegraphics[width=1.0\columnwidth]{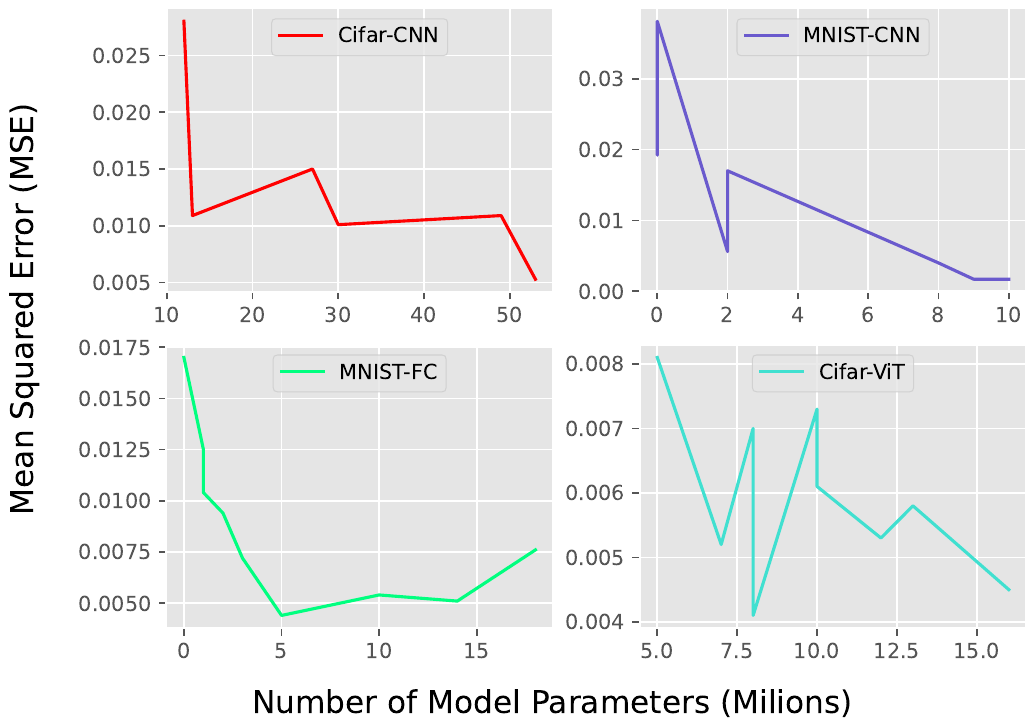}
    \vspace{-1.5em}
    \caption{Plots showing the relationship between the number of parameters and a model's memorization capability.}
    \label{fig:Model_sizeVsMSE}
        \vspace{-1em}
\end{figure}

\textbf{Results.} 
The results presented in Table \ref{table:architecture_size} show how the width and depth impact a model's memorization capacity. As can be seen, both hyperparameters improve pixel accuracy. However, the model width tends to play a greater role. This indicates that for the task of memorization, compression of $\mathcal{D}$ into $\theta^T$ is limited by how many features the samples have in common. As a result, improved memorization capacity can be achieved by increasing the number of neurons per layer as opposed to increasing the total number of layers. 
Fig. \ref{fig:Model_sizeVsMSE} plots the performance of the models as a function of their parameter counts. The plot shows that increasing the number of parameters does not improve memorization capacity. Rather, it is better to use wider and shallower networks, as shown in Fig. \ref{table:architecture_size}. Finally, we note that CNNs require significantly more model parameters to memorize data than other architectures. This may be due to a conflict between $f$ and $f'$ in how the weights are optimized for each task.  

In summary, it appears that it is better to use wider models than deeper models for memorization. However, this will likely depend on how well the dataset can be compressed as a hierarchy of features. 

\begin{table}[t]
\caption{The influence of model depth and model width on a transpose model's memorization ability. Performance is measured in average MSE. Best performances are in bold.}
\vspace{-1em}
\label{table:architecture_size}\vspace{1em}
\centering

\begin{tabular}{c}
\resizebox{0.48\columnwidth}{!}{%
\setlength\tabcolsep{4 pt}
\begin{tabular}{cccc}
\multicolumn{4}{c}{\texttt{MNIST-FC} (30K samples)}              \\ \hline\hline
& \multicolumn{3}{c}{Number of Layers} \\ \hline
FC DIM    & 2           & 3          & 4         \\ \hline
512        & 0.0170      & 0.0125     & 0.0104    \\
1024       & 0.0094      & 0.0072     & \textbf{0.0044}    \\
2048       & 0.0054      & 0.0051     & 0.0076    \\\hline\hline
&&&
\end{tabular}}

\resizebox{0.48\columnwidth}{!}{%
\setlength\tabcolsep{4 pt}
\begin{tabular}{cccc}
\multicolumn{4}{c}{\texttt{MNIST-CNN} (4096 samples)}        \\ \hline\hline
             & \multicolumn{3}{c}{Number of Layers} \\ \hline
\#channels   & 2           & 3          & 4         \\ \hline
64           & 0.0201      & 0.0192     & 0.0381    \\
128          & 0.0056      & 0.0038     & 0.017     \\
256          & \textbf{0.0017 }     & \textbf{0.0017}     & 0.004     \\ \hline\hline
&&&
\end{tabular}}\\
% \caption{Bar}
% }

\resizebox{0.48\columnwidth}{!}{%
\setlength\tabcolsep{4 pt}
\begin{tabular}{cccc}
\multicolumn{4}{c}{\texttt{CIFAR-CNN} (1024 samples)}              \\ \hline\hline
             & \multicolumn{3}{c}{Number of Layers} \\ \hline
\#Channels & 2                 &  3          & 4        \\ \hline
256        &  0.0109           &  0.028        & 0.0560          \\
384        &  0.0101           &  0.015         & 0.0510        \\
512        &  \textbf{0.0081}           &  0.0109          & 0.0473           \\ \hline\hline

\end{tabular}}
% \caption{Foo}

% \hfill
\resizebox{0.48\columnwidth}{!}{%
\setlength\tabcolsep{4 pt}
\begin{tabular}{cccc}
\multicolumn{4}{c}{\texttt{CIFAR-ViT} (4096 samples)}      \\ \hline\hline
           & \multicolumn{3}{c}{Number of Layers} \\ \hline
MLP Dim    & 5           & 7          & 9         \\ \hline
384x2      & 0.0081      & 0.007      & 0.0073    \\
384x3      & 0.0052      & 0.0061     & 0.0051    \\
384x4      & \textbf{0.0041}      & 0.0053     & 0.0043    \\ \hline\hline

\end{tabular}}
% \caption{Bar}

\end{tabular}

\end{table}

\subsection{Ablation Study}\label{subsec:ablation}
\begin{table}[t]
\caption{An ablation study for the proposed spatial index. N and NG  stand for n-ary code and n-ary Gray-code respectively. R stands for random class embeddings. The highlighted rows are the indexing methods used in this paper. }\label{tab:ablation}
\vspace{1em}

\resizebox{\columnwidth}{!}{%
\begin{tabular}{cccccccc}
%\multicolumn{8}{c}{Ablation Study Spatial Index}                                                                                                                                                                                                                                                                                                                            \\ \hline\hline
\multicolumn{2}{c}{\begin{tabular}[c]{@{}c@{}}Sample\\ Enumertion\end{tabular}}                & \multicolumn{2}{|c}{\begin{tabular}[c]{@{}c@{}}Class\\  Emebedding\end{tabular}}                					 & \multicolumn{2}{|c}{\begin{tabular}[c]{@{}c@{}} \texttt{MNIST-CNN}\\ 10K Samples \end{tabular}} 		& \multicolumn{2}{c}{\begin{tabular}[c]{@{}c@{}}\texttt{CIFAR-ViT}\\ 3K Samples\end{tabular}} \\ \hline\hline
\multicolumn{1}{c|}{N}                        & \multicolumn{1}{c|}{NG}                         & \multicolumn{1}{c|}{N-Hot}                         & \multicolumn{1}{c|}{R}                    				 	   & $f$ ACC             																				 & $f'$ MSE                                               & $f$ ACC                             & $f'$ MSE                                                          \\ \hline
\multicolumn{1}{c|}{$\bullet$}                        & \multicolumn{1}{c|}{}                          & \multicolumn{1}{c|}{}                          & \multicolumn{1}{c|}{}                          			                               & \textbf{98.16}                  & 0.026                & \textbf{84.37}																					 & 0.005                                                           \\
\multicolumn{1}{c|}{}                         & \multicolumn{1}{c|}{$\bullet$}                         & \multicolumn{1}{c|}{}                          & \multicolumn{1}{c|}{}                          			                             & 97.73                           & 0.024                  & 83.27       																					   & 0.005                                                           \\
\rowcolor[HTML]{EFEFEF}                                                                                                                                                                                                                                                                                                                                                         
\multicolumn{1}{c|}{\cellcolor[HTML]{EFEFEF}} & \multicolumn{1}{c|}{\cellcolor[HTML]{EFEFEF}$\bullet$} & \multicolumn{1}{c|}{\cellcolor[HTML]{EFEFEF}$\bullet$}  & \multicolumn{1}{c|}{\cellcolor[HTML]{EFEFEF}} 	                               & 98.00                           & \textbf{0.014}  		          & 82.4  																					 & 0.003                 				\\
 \rowcolor[HTML]{EFEFEF}                                                                                                                                                                                                                                                                                  
\multicolumn{1}{c|}{\cellcolor[HTML]{EFEFEF}} & \multicolumn{1}{c|}{\cellcolor[HTML]{EFEFEF}$\bullet$} & \multicolumn{1}{c|}{\cellcolor[HTML]{EFEFEF}} & \multicolumn{1}{c|}{\cellcolor[HTML]{EFEFEF}$\bullet$}  	                              & 97.88                           & 0.021                         & 81.08  																					& \textbf{0.0024}                                                
                                                                                                                                                                                                                                                                 
                                              \\ \hline\hline
\end{tabular}}

\end{table}

\textbf{Setup.} In Section \ref{sec:mem} we described the spatial indexer used to teach a model how to store and retrieve specific samples from $\theta^T$. To evaluate the contribution of using Gray code and class-based projections, we performed an ablation study on the spatial indexing function. 

\textbf{Results.} 
First we compared different ways to systematically map natural numbers to an $n$-dimensional space. We experimented with $n$-ary codes and $n$-ary Gray codes for $n=3$. The results presented in Table \ref{tab:ablation} show that although the primary task $f$ performs better when the attacker uses $n$-ary codes, the performance of $f'$ improves when $n$-ary Gray codes are used. The performance of $f'$ is improved further when $E(c)$ is added to project samples into different sub-spaces according to their class. Both n-hot encodings and random embedding significantly improve memorization performance. However, we found that FC and CNN models perform better with n-hot embeddings, and ViT models perform better with random embeddings.

%-------------------------------------------------------------------------------
\section{Countermeasures}\label{sec:counter}
%-------------------------------------------------------------------------------

In general, the most effective approach to preventing a transpose model attack is to analyze the training code before executing it in a protected environment. One can also try to  mitigate the attack by fine-tuning the trained model on the primary task or force the platform users to incorporate some form of weight regularization.
However, these approaches requires experimentation since using the wrong optimization settings can harm a trained model, and thus reduce the usability of the platform.
We provide experimental results for using fine-tuning and weight regularization as countermeasures in Appendix~\ref{appx:prevention}.
Our results show that fine-tuning only mitigates the attack for CNNs, and that using $L_2$ weight regularization only mitigates memorization in the case of \texttt{MNIST}. This is because a sufficiently large decay factor harms the primary task.

Despite the potential of these approaches, in our threat model, the defender is not the code author (e.g., DTaaS, federated learning, etc.) To use the above approaches, the defender would have to manually reverse-engineer and inspect every users' training code to identify the transpose training segment and remove it. Analysis of training code requires a significant amount of time, resources and technical ability. Detecting the secondary training objective is not trivial either since many models make use of multiple training passes over the same weights \cite{bardes2021vicreg,yan2021agg}, and code can be obfuscated. Therefore, in this section we explore defenses which can be \textit{automated}.

\begin{figure}[t]
\centering

\includegraphics[width=0.49\columnwidth]{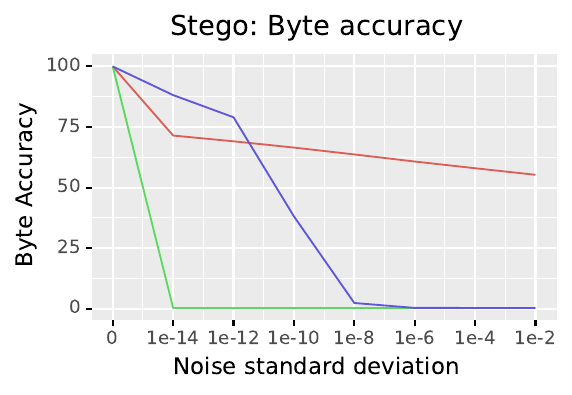}  \includegraphics[width=0.49\columnwidth]{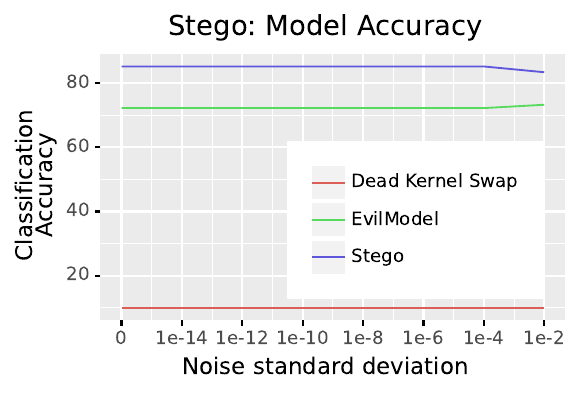} \\
\centering \includegraphics[width=0.49\columnwidth]{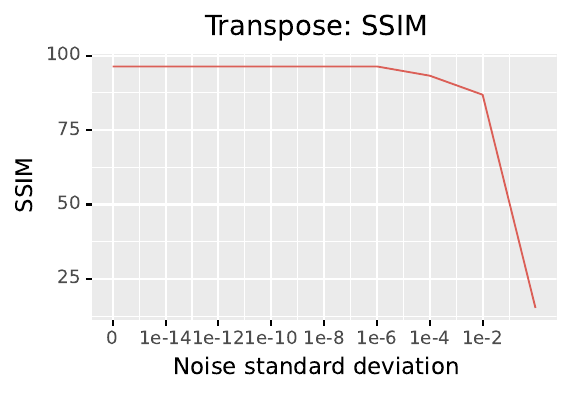}   \includegraphics[width=0.49\columnwidth]{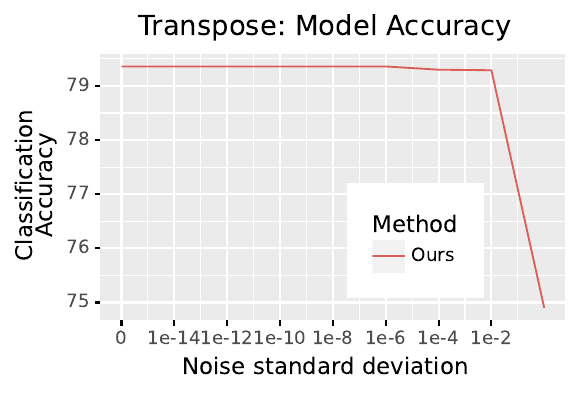} \\
\vspace{.5em}
\includegraphics[width=\columnwidth]{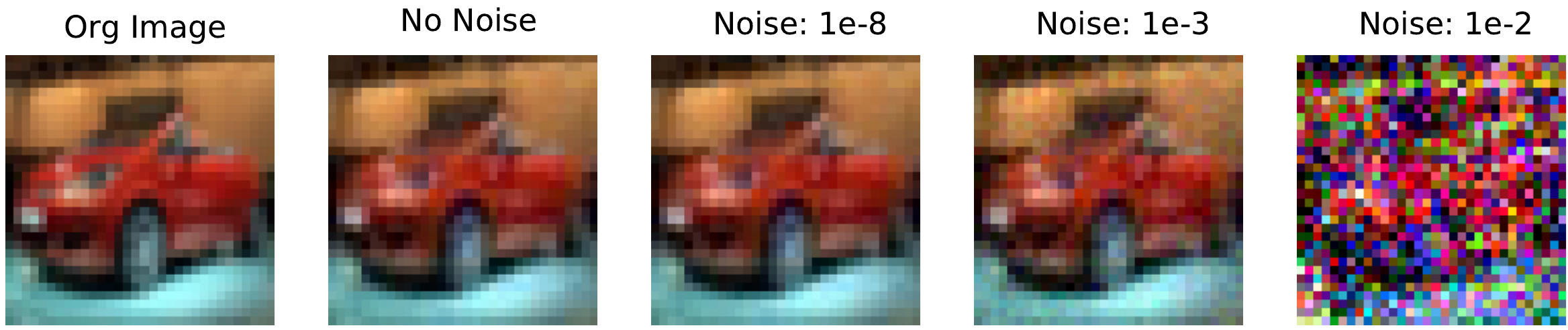}
\vspace{-1em}
\caption{Robustness of various model-based data exfiltration attacks to additive noise on the model's parameters. Top: Noise effect on Stego methods, bottom: Noise effect on Transposed models.}
\label{figure:stego}
\end{figure}

\subsection{Prevention}
To prevent a transpose attack, one may consider adding Gaussian noise to the parameters of a network after training but before their exportation. We found that with enough noise, the secondary task of memorization can be mitigated but at a severe cost to the primary task's performance. 

To evaluate this method, we compare the robustness of a transpose model to other model-based data exfiltration attacks. Specifically we examine three steganographic approaches which hide binary in a model's parameters: (1) \textit{StegoNet} which stores data in the least significant bytes (LSBs) \cite{liu2020stegonet}, (2) \textit{EvilModel} which stores data in the last three bytes \cite{wang2021evilmodel}, and (3) a baseline we call Dead Kernel Swap which simply replaces low $L_1$ norm neurons with the image data. 
In our experiment, we trained a ViT model as a classifier on the CIFAR-10 dataset and we used all four methods to hide 1,024 CIFAR-10 images inside it. 

Fig. \ref{figure:stego} presents the results of this experiment. The results show that only $1e^{-8} \sigma$ of additive noise is required to mitigate steganographic approaches while maintaining the performance of the primary task. In contrast, adding the same level of noise to a transpose model hardly affects the secondary task. This is because the parameter noise directly affects the LSBs but not the abstract concepts learnt by the model. We also note that although the SSIM drops significantly at 1e-3 noise for the transpose attack, the images are still recognizable.

\subsection{Detection}
We propose one possible way to detect transpose memorization models. Our approach is inspired by the work of \cite{shan2020gotta} which suggests placing gradient honeypots in $\theta$ to detect adversarial examples. We suggest that if $f'_{\theta^T}$ is not a transpose model trained to perform memorization, then it will be extremely hard to have $f'_{\theta^T}$ generate content that resembles the distribution of $\mathcal{D}$. On the other hand, if $f'_{\theta^T}$  has been memorizing data, then it should be possible to optimize an input code $e$ such that $f'_{\theta^T}(e)$ produces content. 

The algorithm is similar to generating an adversarial example with no bound epsilon and a target loss that decreases the output's entropy. In our implementation, we first initialized $e^{(0)} \sim \mathcal{U}(0, 1)$. We then performed the basic iterative method (BIM \cite{kurakin2018adversarial}) until convergence
%\vspace{-.5em}
\begin{equation}\label{eq:bim}
    e^{(i+1)} = e^{(i)} - \alpha \cdot \nabla_e \mathcal{L}_2 \left(f'_{\theta^T}(e^{(i)}),\bar{x}\right)
\end{equation}
where $\bar{x} = \frac{1}{m} \sum_i x_i$ for $x_i \in \mathcal{D}$ and $\mathcal{L}_2$ is the standard $L_2$ loss. After convergence (or $k$ iterations), we compute the MSE score $\mathcal{L}_2(e^{(k)},\bar{x})$. If the score is above a predetermined threshold, then we flag the model as malicious. To improve accuracy, this approach can be repeated a number of times with different random starts; the lowest score is then selected. In the Appendix, we suggest a method for selecting the threshold without prior knowledge of $f'_{\theta^T}$.

To evaluate this approach, we performed 20 trials on both the benign and malicious (transposed) versions of \texttt{MNIST-FC}, \texttt{MNIST-CNN}, \texttt{CIFAR-CNN}, \texttt{CIFAR-ViT}, and \texttt{CelebA-ViT}, where the benign versions of these models are simply $f$ with no secondary task in $\theta^T$. In each trial, we performed 300 iterations of equation (\ref{eq:bim}). We achieved an area under curve (AUC)\footnote{An AUC of 1.0 indicates a perfect classifier, and an AUC of 0.5 indicates that the model is guessing randomly.} score of 1.0 for all models except \texttt{MNIST-FC} which achieved an AUC of 0.95. Table \ref{tab:defense} provides statistics on the scores obtained in the trials. The table shows that there is a significant gap between the scores obtained from benign and malicious models.

In summary, the proposed countermeasure is effective. It is also has an advantage; it does not make any assumptions regarding the architecture or indexing strategy the adversary might choose. However, there are several disadvantages: (1) it assumes that $f'_{\theta^T}$ will be trained to memorize many samples, (2) the defender will have to design a framework which can transpose arbitrary models, and (3) execution of this algorithm requires additional resources (GPU acceleration etc.) We encourage the research community to look into better ways of preventing transpose attacks against data exfiltration and other potential secondary tasks.

%-------------------------------------------------------------------------------
\section{Related Work}\label{sec:relworks}
\begin{figure}[t]
    \centering
    \includegraphics[width=\columnwidth]{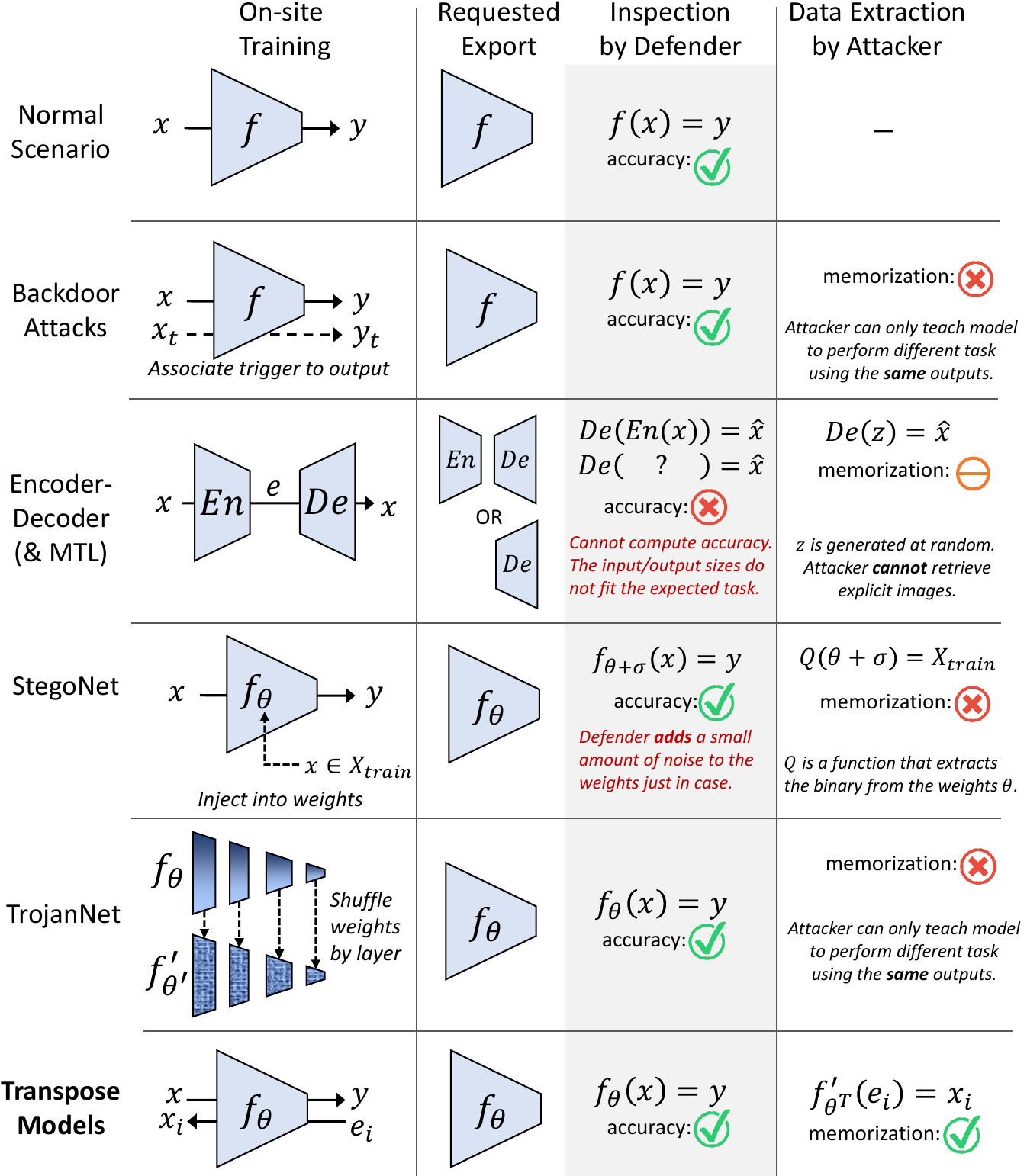}
        \vspace{-2em}
    \caption{A comparison of existing hidden model and \textit{intentional} memorization attacks. The columns (left to right) show the order of events during an attack. The inspection step indicates whether the attack is covert during export, and the data extraction step indicates whether the covert model can be used to memorize explicit images.}
    \label{fig:contrast}
\end{figure}

%-------------------------------------------------------------------------------
In this paper we introduce novel methods for implementing hidden models (the transpose attack) and data exfiltration (memorizing data via spatial indexing). In this section we review the state-of-the-art in both domains. In Fig. \ref{fig:contrast}, we compare the transpose attack to known hidden model attacks and other \textit{intentional} memorization attacks.

\subsection{Hidden Models}

The transpose attack is similar to MTL where a single neural network is trained to perform several different tasks \cite{zhang2021survey}. In MTL, all tasks are passed through the network in the forward direction, and each task receives dedicated layers at the output (heads) to make predictions. Classic MTL is not covert, since it requires additional heads which may be considered suspicious during export.

To be more covert, a model can learn a hidden secondary task with the same weights used for the primary (overt) task. One instance of this kind of attack is the backdoor attack. In a backdoor attack, a model $f$ is conditioned during training to produce a specific output if a specific trigger pattern is presented in the input \cite{li2022backdoor}. This attack can be used to allow models to exfiltrate knowledge obtained from $\mathcal{D}_{train}$. For example, in \cite{bagdasaryan2021blind}, the authors showed how a model can be trained to count faces in an image but then output an individual's identity when a trigger is presented.

\begin{table}[t]
\caption{The mean and standard deviation of the detection scores obtained from each of the models.}\label{tab:defense}
\vspace{1em}
\centering
\begin{tabular}{lll}
\hline\hline
          & Benign   & Transposed  \\ \hline
\texttt{MNIST-FC}  & 0.031$\pm$0.0   & 0.007$\pm$0.010 \\
\texttt{MNIST-CNN} & 0.025$\pm$0.0   & 0.012$\pm$0.002 
\\
\texttt{CIFAR-CNN} & 0.0149$\pm$0.0  & 0.007$\pm$0.002 \\
\texttt{CIFAR-ViT} & 0.226$\pm$0.007 & 0.002$\pm$0.005 \\
\texttt{CelebA-ViT} & 3.596 $\pm$0.615   & 0.002$\pm$0.0
\\ \hline\hline
\end{tabular}
\end{table}

If the attacker does not need to trigger the secondary task (e.g., will have white-box access to the model after export), then the secondary task can be embedded as a model within $\theta$. For example, in \cite{guo2020trojannet}, the authors proposed \textit{TrojanNet}. While training a primary model $f$ on $\theta=\{\theta_0,\theta_1,\theta_3,...\}$ a secondary model $f'$ is trained on $\theta'=\{\theta'_0,\theta'_1,\theta'_3,...\}$ in tandem. In their work, they showed that the weights from the $i$-th layer of $f'$ can be a random permutation of the $i$-th layer's weights in $f$. In other words, if $\theta'_i = \text{shuffle}(\theta_i)$. Then an attacker  who knows the mapping can extract $f'_\theta$ from $\theta$.

We make two distinctions between existing hidden model attacks and transpose attacks. First, in \cite{guo2020trojannet}, the model $f'$ must preserve the same network architecture as $f$. This means that if the expected primary task is image classification, then the secondary task is \textbf{limited} to tasks with the same input and output sizes. For example, if $f$ performs cancer detection on CT images of size 512x512, then $f'$ must take inputs of size 512x512 and produce outputs of size 2. This would prohibit the task of sample memorization. Second, transpose models present a novel vulnerability, since no one has yet considered that a model can be executed in reverse. Modern defenses detect backdoor and Trojan models by analyzing a model's response to various inputs~\cite{chen2018detecting, tran2018spectral, guo2020trojannet, Dong_2021_ICCV}. However, this is done in the forward direction which would overlook the transpose direction we have proposed.

\subsection{Data Extraction}
As discussed in Section \ref{sec:intro}, neural networks can either implicitly or explicitly be taught to memorize samples from a dataset $\mathcal{D}$, where $\mathcal{D}$ is $\mathcal{D}_{train}$ or some other dataset. When $\mathcal{D}=\mathcal{D}_{train}$, gradients and other signals from $f_\theta$ can be used to extract samples which reflect $\mathcal{D}_{train}$ ~\cite{fredrikson2015model,zhang2020secret}. However, these approaches are opportunistic, since the attacker has no control over which samples will be memorized, and in some cases only blurry approximations can be extracted \cite{yang2019neural}.

In \cite{tramer2022truth}, the authors showed that an attacker with access to $\mathcal{D}_{train}$ can poison the dataset to cause the model to memorize samples better, leading to improved data extraction. Moreover, in \cite{haim2022reconstructing}, it was shown that for very small networks trained on small datasets, it is possible to extract key samples implicitly memorized by the model by solving a system of equations. However, in both of these cases, the attacker cannot choose which samples to memorize nor systematically retrieve all of the memorized samples from the model.

Some studies have shown that it is possible to memorize selected samples from $\mathcal{D}$. For example, in \cite{li2022data}, the authors used MTL to train a model to perform both medical image segmentation and image reconstruction using two separate heads. Then, after export, the images memorized by the second head are retrieved by using the input encodings. This approach is not covert, since the secondary task is apparent in the model's architecture and the encodings must be exported with the model. Furthermore, the approach does not enable the attacker to systematically retrieve all of the memorized samples without the encodings. 

Instead of embedding the data in the model's function, other studies tried embedding the data in the model's parameters using stenography \cite{liu2020stegonet,wang2021evilmodel,hitaj2022maleficnet}. 
However, as shown in Section \ref{sec:counter}, we found that these methods are easily mitigated during export if a small amount of random Gaussian noise is added to the parameters. In contrast, significantly larger amounts of additive noise are required to affect a transpose model.
Regardless, the memorization technique proposed in this paper provides a novel approach for data exfiltration, which if overlooked, can be used by attackers to perform data exfiltration attacks undetected.

\section{Conclusion}
In this paper we introduced two novel concepts. The first is the transpose attack, in which a network is trained to perform a secondary task where the task is hidden since it can only be executed by flipping (transposing) the network. The second is the task of sample memorization. We achieve this task by developing a spatial indexer that enables users to retrieve specific samples from trained models. By putting these concepts together, an attacker can extract data from protected environments through an attack vector which is currently being overlooked. Through our evaluations, we showed that this attack can not only be used to violate the confidentiality of protected datasets and steal intellectual property by utilizing the stolen data off-site. Finally, we suggested one possible way to detect this attack. We hope that his work help bring awareness to this new class of attacks and encourages researchers to find ways to mitigate it.

\section*{Acknowledgment}
\noindent
This paper received funding from European Union's Horizon 2020 research and innovation program under grant agreement 952172 and support from the Zuckerman STEM Leadership Program.
\begin{figure}[h]
    \centering
     \includegraphics[width=.1\textwidth]{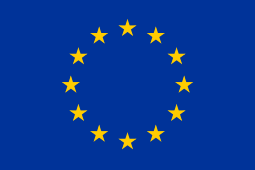}

\end{figure}


\bibliographystyle{unsrt}
\bibliography{paper}

\appendices
% \section*{Appendix}
\section{Additional Results}
\subsection{Examples of other Secondary Tasks}\label{appx:othersecondary}
Aside from performing memorization for the purpose of data exfiltration, the secondary task can be used to exploit proprietary data or reveal confidential information in the provided dataset, or other exposed datasets. The following are some examples of other secondary tasks which can potentially be performed using transposed models:
\begin{enumerate}
    \item The secondary task is used to perform a sensitive operation such as predicting the identity of faces in an image where the primary task performs a permitted non-sensitive task such as counting faces in an image. This attack scenario has been suggested in previous works such as \cite{bagdasaryan2021blind}.
    
    \item The secondary task is used to reveal confidential statistics about the dataset (gender balance, average salaries, ...). This attack is similar to property inference attacks~\cite{xu2020subject, parisot2021property}. However, here $f'$ would be used to explicitly reveal information about $\mathcal{D}_{train}$ as opposed to implicitly revealing this information by analyzing $f$.  
    % property inference
    \item The secondary task is used to explicitly perform membership inference (predicts if a set of attributes belong to a sample in the dataset). This is similar to the final classifier used in a shadow model attack except that here there is no need to train surrogate models to identify membership via confidence vectors ~\cite{shokri2017membership}.
    % shadow models
    
    \item The secondary task is used to memorize something other
    than a dataset, such as auxiliary information. For ex-
    ample, medical imagery datasets are stored as DICOM
    files and often include confidential information, such
    as a patient’s date of birth or country of residence~\footnote{https://www.dicomlibrary.com/dicom/dicom-tags/}.
    Attacks targeting hospitals' DICOM databases have been widely studied
    and pose a significant breach of privacy~\cite{desjardins2020dicom}.
    Using the transpose model, an adversary can train $f$ for a benign medical task (e.g. lesion detection), while simultaneously training $f'$ to memorize
    confidential DICOM attributes.
    % For example, medical datasets are usually stored as  DICOM files and are also often accompanied by an annotation file. DICOM files and annotation files both contain sensitive information.
    % DICOM files contain metadata such as a patient's date of birth and country of residence\footnote{https://www.dicomlibrary.com/dicom/dicom-tags/} and annotation files, as in the case of the QIN-HEADNECK dataset~\footnote{https://wiki.cancerimagingarchive.net/display/Public/QIN-HEADNECK}, can contain information about the patient's age, weight and ethnicity.
    
    \item The secondary task uses auxiliary data to predict sensitive user attributes given non-sensitive ones. 
    As in Model Inversion attacks~\cite{veale2018algorithms}, the hidden model can serve as a predictor for sensitive attributes available in the protected environment such as the ethnicity of a patient in the case of medical data.
    % model inversion
    
    \item The secondary task is used to perform a non-licensed task, such as using the dataset to train a generator that creates similar images. Other works also explore the potential of adversaries training models to perform restricted tasks on protected data (e.g., \cite{li2022data}). Here, we suggest transposed models can do this more covertly and potentially on very different tasks as well.
\end{enumerate}

\subsection{Prevention: Complete Evaluation}
\label{appx:prevention}
In this section we provide the complete experiment performed to evaluate the impact of fine-tuning and regularization as a defence against transpose attacks. In all of our experiments we trained five transposed architectures MNIST-FC, MNIST-CNN, CIFAR-CNN, CIFAR-ViT, CelebA which memorized 30k, 10k, 1k, 2k and ~5k\footnote{200 identities} samples respectively.

\textbf{Fine-tuning:}
For the purpose of this experiment, we assume that the defender will simply copy the optimization settings from the training code to perform the fine-tuning (e.g., hyper-parameters, learning rate schedulers, etc.) This is a practical assumption since the defender is not the code author. We evaluated fine-tuning on all of the architectures and datasets used in this paper.

\begin{figure*}
    \centering
    \begin{tabular}{cc}
    \includegraphics[scale=0.5]{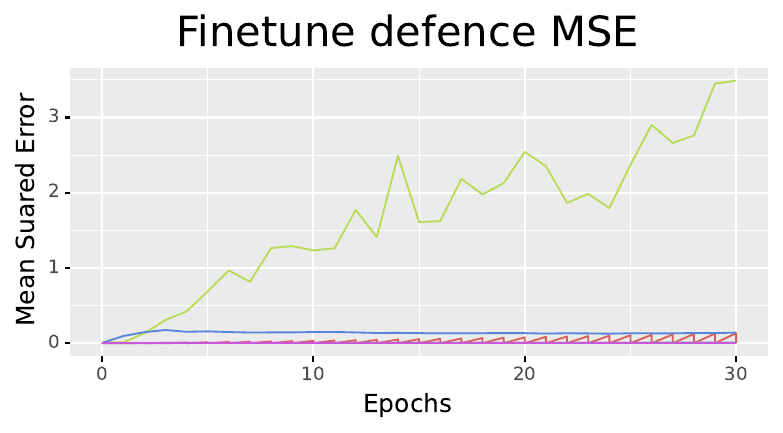} & \includegraphics[scale=0.5]{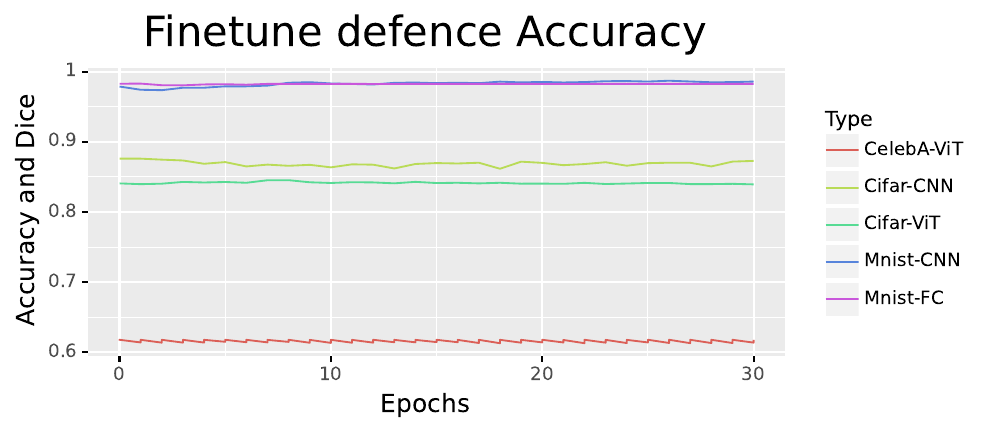} \\
    \end{tabular}
    \vspace{1pt}
    \caption{Fine-tuning the model only on the primary task as a defense}
    \label{fig:finetuning}
\end{figure*}

We found that for all models and datasets fine-tuning the primary task did not affect the secondary task of model memorization (see Fig. ~\ref{fig:finetuning}). The exception is for CNN architectures where five epochs of fine-tuning significantly harmed the secondary task. We believe the reason why CNNs are more sensitive to fine-tuning than FC and ViT is that convolution weights are less flexible in terms of multi-tasking and because these layers contain fewer parameters. As a result, when fine-tuning a primary task in a CNN the weights relating to the secondary task are significantly impacted. 

We note that although fine-tuning a CNN appears to be an effective countermeasure, this approach cannot be automated in most cases. This is because the defender must analyze the training code to execute training on the primary task only. In cases where complete automation of the defence is required, our proposed detection algorithm can be used instead.

\textbf{Regularization: Weight Decay}
\begin{table}[]
\caption{The primary task accuracy/MSE of models trained with a weight decay of $\lambda$}\label{tab:defense}
\vspace{1em}
\centering
% \DIFaddbeginFL
\begin{tabular}{cccc}
\hline\hline
           & $\lambda$=0.1 & $\lambda$=0.01 & $\lambda$=0.001 \\ \hline
\texttt{MNIST-FC}   & 97.78 / 0.0113 & 97.77 / 0.0053 & 98.17 / 0.0058  \\
\texttt{MNIST-CNN}  & 98.17 / 0.0141 & 97.84 / 0.0083 & 98.15 / 0.0079  \\
\texttt{CIFAR-CNN}  & 68.84 / 0.019  & 87.86 / 0.0113 & 87.29 / 0.0137  \\
\texttt{CIFAR-ViT}  & 66.04 / 0.0043 & 82.35 / 0.0024 & 82.55 / 0.0017  \\
\texttt{CelebA-ViT} & 64.28 / 0.0047 & 63.04 / 0.0047 & 62.32 / 0.0047  \\ \hline\hline
\end{tabular}
% \DIFaddendFL
\label{tab:weight_dacay}
\end{table}
We used $L_2$ regularization using the AdamW optimizer in Torch. We evaluated the impact of this regularization on all of the datasets and architectures used in this paper.

We found that for all cases, with the exception of MNIST, the performance of the secondary task was not affected by the regularization (see Table~\ref{tab:weight_dacay}). Therefore, it appears that weight decay is not an effective defence for models trained on datasets which are more complex than MNIST.

\subsection{Countermeasure: Threshold Selection}
\begin{figure*}[t]
    \centering
    \includegraphics[width=2.1\columnwidth]{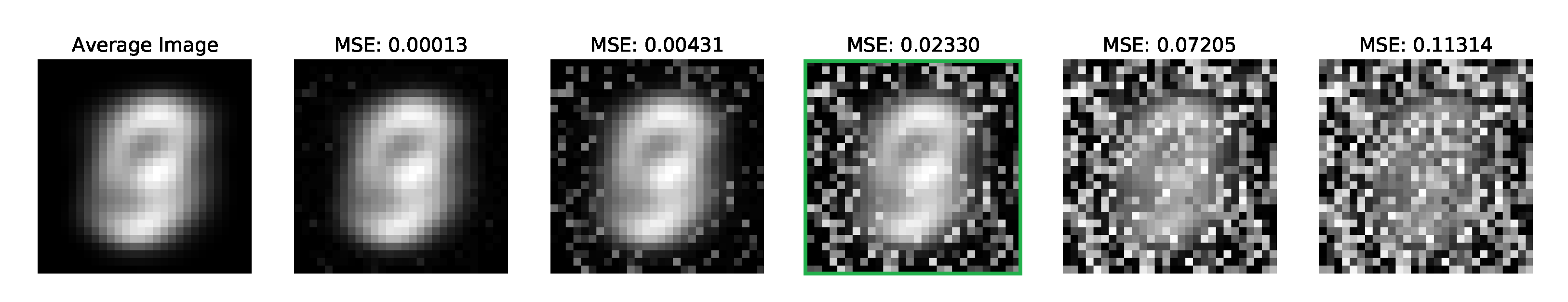}
    \caption{A visualization for the subjective method of selecting a good threshold. Left to right: the average image $\bar{x}$ with increasing level of noise to it. The sample in green shows where the content start degrading, and its MSE value is selected as the threshold (MSE between that sample and the clean $\bar{x}$).}
    \label{fig:counter_measure_threshold}
\end{figure*}

It is possible to define a suitable threshold in advance, without knowledge of $f'_{\theta^T}$. One way is to add AWGN to $\bar{x}$ until the content is subjectively no longer visible. Then the threshold can be set to the MSE of the noisy sample and $\bar{x}$. Fig. \ref{fig:counter_measure_threshold} visualizes this process on the MNIST dataset. With an MSE of approximately 0.02330 the noisy version of $\bar{x}$ loses its integrity. Therefore, we use 0.02330 as our threshold. 

Using this method on our models (see Table \ref{tab:defense}) we obtained true positive and false positive rates of 1.0 and 0.0 respectively for \texttt{MNIST-CNN}, \texttt{CIFAR-CNN},  \texttt{CIFAR-ViT}, and \texttt{CelebA-ViT}. On \texttt{MNIST-FC} we obtained a true positive rate of 1.0 and a false positive rate of 0.1.

\section{Artifact Appendix}

\subsection{Description \& Requirements}

In this artifact we present information on where to obtain python code for creating your own transpose models for memorization attacks. The current version of the code supports fully connected (FC) neural networks and will be extended to CNNs and vision transformer networks in the near future.
\textbf{DOI: }\href{https://zenodo.org/badge/latestdoi/684759687}{https://zenodo.org/badge/latestdoi/684759687}

\subsubsection{How to access}
We have uploaded the code to a GitHub repository, which can be accessed via this link: \\
\href{https://github.com/guyAmit/Transpose-Attack-paper-NDSS24-/tree/main
}{https://github.com/guyAmit/Transpose-Attack-paper-NDSS24-/tree/main}\\
We also supply a self-contained Colab notebook, which allows running the demo without the need to install anything:
\href{https://colab.research.google.com/drive/1iFoKCheq3UZLdPxRj0SkqvRnkUsvc-Ia?usp=sharing
}{https://colab.research.google.com/\\drive/1iFoKCheq3UZLdPxRj0SkqvRnkUsvc-Ia?usp=sharing}.
\\

% Describe here how to access your artifact. During the artifact evaluation, in case of a private repository, you should provide instructions on how to access it. For the camera-ready version of the appendix, you must provide a DOI link to the AEC-approved artifact version that you must upload by then to permanent storage.

\subsubsection{Hardware dependencies}
The minimal hardware requirement for running the code locally is a CPU with at least 8GB of RAM. 
We do recommend using a machine containing a GPU such as Nvidia-RTX1080 for convenience.
If neither are available, check out our Colab demo(link above).

\subsubsection{Software dependencies} 
% Simply write ``None." where this does not apply to your artifact.
A full list of software packages is provided in section~\ref{sec:Installation}.

% \subsubsection{Benchmarks} 
% % Describe here any data (e.g., datasets, models, workloads, etc.) required by the experiments with this artifact reported in your paper. Simply write ``None." where this does not apply to your artifact.
% \begin{itemize}
%     \item The dataset used in the artifact is supplied via the torchvision package(installation below), and there is no need to download the dataset independently
%     \item The code for the FC neural network used in the demo is in the $model.py$ file within the $src$ directory.
% \end{itemize}

\subsection{Artifact Installation \& Configuration}
\label{sec:Installation}
% This section should include all the high-level installation and configuration steps required to prepare the environment to be used for the evaluation of your artifact.
All of our experiments ran on an Anaconda environment with access to a GPU.
The required Python packages for the demo are provided below:\\
\begin{enumerate}
    \item numpy=1.19.2
    \item jupyterlab=3.2.5
    \item pytorch=1.8.1
    \item torchvision=0.9.1
    \item scipy=1.4.1
    \item scikit-image=0.17.2
    \item scikit-learn=0.22.1
    \item matplotlib=3.2.2=1
\end{enumerate}

To run the code, create a new Anaconda environment with the listed packages, and start JupyterLab.
Using JupyterLab, open the '$Example.ipynb$' file and follow the instructions within the notebook.

\subsection{Experiment Workflow}
% This section should provide a high-level view of the experimental workflow and how it is implemented, invoked, and (if needed) customized. The section is optional if the experiment workflow can be easily embedded in the Evaluation section.
In the current version of the artifact, we supply a demo of the Transposed attack.
The demo includes training a transpose model on the MNIST dataset and the extraction process of the memorized images.
In each run of the demo, the user can adjust the experiment parameters e.g. percentage of memorized samples, and examine the effect on the extracted images.

\subsection{Customization}
% Provide here notes on how to customize your experiments, when applicable. The section is optional.

In the '$Example.ipynb$' notebook under the title "Run Parameters" is a cell that enables setting the experiment parameters.
Before running the notebook, set the parameters to desired values, such as:
\begin{itemize}
    \item $input\_size$ = 784
    \item $output\_size$ = 10 (number of classes in classification)
    \item $hidden\_layers$ = [1024, 1024, 1024]
    \item $percentage\_to\_memorize$ = 0.1  
    \item $batch\_size$ = 128
    \item $epochs = 200$
    \item $save\_path = './models/mnist\_example.ckpt'$
\end{itemize}

The duration of the training process will vary depending on the hardware, model layers size, and the number of memorized images.
The configuration set in the notebook takes about 20 minutes to run using our Nvidia-RTX3090 GPU.

\subsection{Notes}
% This section is meant to allow authors to include any further important notes that may not fall within any of the previous categories. We kindly encourage you to remove this section where this sort of content may not be strictly needed (rather than filling it with unnecessary or redundant information)
The current version only supports fully connected (FC) neural networks and comes with some helper classes for demonstrating the attack with the MNIST handwritten digit dataset. In the coming months, we will integrate into the library CNNs and Vision Transformers, but for the time being, we supply two notebooks demonstrating how to train transpose Vision Transformers and CNNs on the Cifar dataset.

Note that the provided code can be used to train transpose models on other datasets.
This can be achieved by adjusting the classes in $dataset.py$ file to fit the new dataset.

\end{document}